\documentclass{article}

\PassOptionsToPackage{numbers,compress}{natbib}

\usepackage[main, final]{neurips_2026}

\usepackage[utf8]{inputenc} %
\usepackage[T1]{fontenc}    %
\usepackage{hyperref}       %
\hypersetup{
  colorlinks=true,
  linkcolor=red,
  citecolor=blue,
  urlcolor=blue,
  hypertexnames=false
}
\usepackage{url}            %
\usepackage{booktabs}       %
\usepackage{amsfonts}       %
\usepackage{nicefrac}       %
\usepackage{microtype}      %
\usepackage{xcolor}         %
\usepackage{graphicx}
\usepackage{subcaption}
\usepackage{amsmath}
\usepackage[most]{tcolorbox}
\tcbset{
  promptbox/.style={
    colback=gray!8,
    colframe=gray!40,
    fonttitle=\bfseries\small,
    title=#1,
    boxrule=0.5pt,
    arc=3pt,
    left=6pt, right=6pt, top=4pt, bottom=4pt,
    before upper={\small\ttfamily},
  }
}
\usepackage{multirow}
\usepackage{enumitem}
\usepackage{xcolor}
\usepackage{listings}
\title{Toward Executable Multi-framework Front-end Code Generation with Self-Correction}

\author{%
  Linxiao Li$^{1*}$ \quad
  Haoran Ma$^{2*}$ \quad
  Chenyue Wang$^{1}$ \\[2pt]
  \bfseries
  Jiaye Lin$^{3}$ \quad
  Haochen Sui$^{4}$ \quad
  Jiechao Gao$^{5\dagger}$ \\[4pt]
  $^{1}$The University of Sydney \quad
  $^{2}$Harvard University \quad \\
  $^{3}$Meituan \quad
  $^{4}$University of Michigan \quad
  $^{5}$Stanford University
}
\begin{document}

\maketitle
{
\renewcommand{\thefootnote}{\fnsymbol{footnote}}
\footnotetext[1]{Equal contribution.}
\footnotetext[2]{Corresponding author: \texttt{jiechao@stanford.edu}.}
}

\begin{abstract}

Generating codes from UI screenshots has recently benefited from MLLMs, yet existing methods centered on HTML targets and transfer poorly to multi-frameworks such as React, Vue, and Angular. 
However, unlike HTML, which does not entail compilation issues, generating executable code in multi-framework settings is more challenging due to framework-specific differences in syntax and failure modes, as well as the need to maintain global consistency across files.
To address this problem, we study executable multi-framework front-end code generation and propose \textbf{MESCoder}, a tree-structured self-corrective generation framework.
Our method first extracts a structural scaffold from the screenshot and then constructs sub-trees and assembly relations step by step, transforming large project generation into a sequence of localized decisions. 
On top of this scaffold, we introduce a unified Project Agent that continuously revises its own partial project under runtime sandbox feedback and dynamically chooses between local repair and cluster level repair. 
For training, we adopt a two-stage strategy that first initializes the policy offline to learn basic multi-framework generation and repair priors, and then refines it with online reinforcement learning to optimize long horizon executability and final page quality. 
The resulting framework models multi-framework executability as a conditioned project state transition process and improves the stability and quality of complex front end generation through self correction.
Experiments show that our method improves executability, 
while narrowing the visual-fidelity gap to HTML-based generation.
Codes are available at: \href{https://github.com/leo-lilinxiao/MESCoder}{Link}.

\end{abstract}

\section{Introduction}
Recent advances in MLLMs have substantially improved the ability to generate front-end code directly from UI screenshots ~\cite{si2025design2code,laurenccon2024unlocking,yun2024web2code,wu2026mulfcoder}. 
This task requires models to jointly perform visual understanding, structural abstraction, and front-end implementation. 
As a result, generation quality depends not only on recognizing local UI elements, but also on whether the model can transform hierarchical layout and spatial organization into a coherent implementation ~\cite{wu2025mllm,liang2025waffle}.

Existing work has made clear progress on this problem, yet the 
dominant formulation remains largely HTML-centered. Recent efforts 
have scaled HTML-screenshot training corpora~\cite{laurenccon2024unlocking,yun2024web2code,gui2025webcode2m} 
and established benchmarks for measuring how faithfully a generated 
page reproduces its reference~\cite{si2025design2code}. 
On the method side, divide-and-conquer decomposition~\cite{wan2025divide}, 
layout-guided prompting~\cite{wu2025mllm}, and structure-aware 
fine-tuning~\cite{liang2025waffle} further improve element recall and 
spatial fidelity. Common to all of these efforts, however, is a 
shared assumption: the generation target is a single self-contained 
HTML file that renders without compilation and without cross-file 
reasoning---an assumption that does not hold in real front-end 
development, where production interfaces are assembled from multiple 
files within framework-specific project structures and become visible 
only after the entire project successfully compiles and runs.

However, extending screenshot-to-code generation from HTML to multi-frameworks such as React, Vue, and Angular is not merely a matter of changing prompts or rewriting syntax. 
Unlike HTML, multi-framework front-end generation requires models to produce an executable project within the framework-specific program space, where correctness depends not only on surface syntax but also on component boundaries, cross-file dependencies, parent-child assembly, and runtime consistency. 
Existing analysis shows that straightforward prompt transfer across frameworks causes a substantial drop in compilation success and is accompanied by highly framework-specific error modes~\cite{xiao2025designbench}. 
In stricter frameworks, failures are often no longer limited to local token- or template-level mistakes, but instead emerge as engineering-level inconsistencies involving import/export relations, bindings, assembly, and project-level coherence. 
Meanwhile, one-shot screenshot-to-project generation entangles local implementation, global composition, and framework constraints in a single decoding process, making intermediate errors difficult to localize and even harder to correct reliably.

\begin{figure}[t]
\vspace{-0.3em}
\centering
\includegraphics[width=1.0\linewidth]{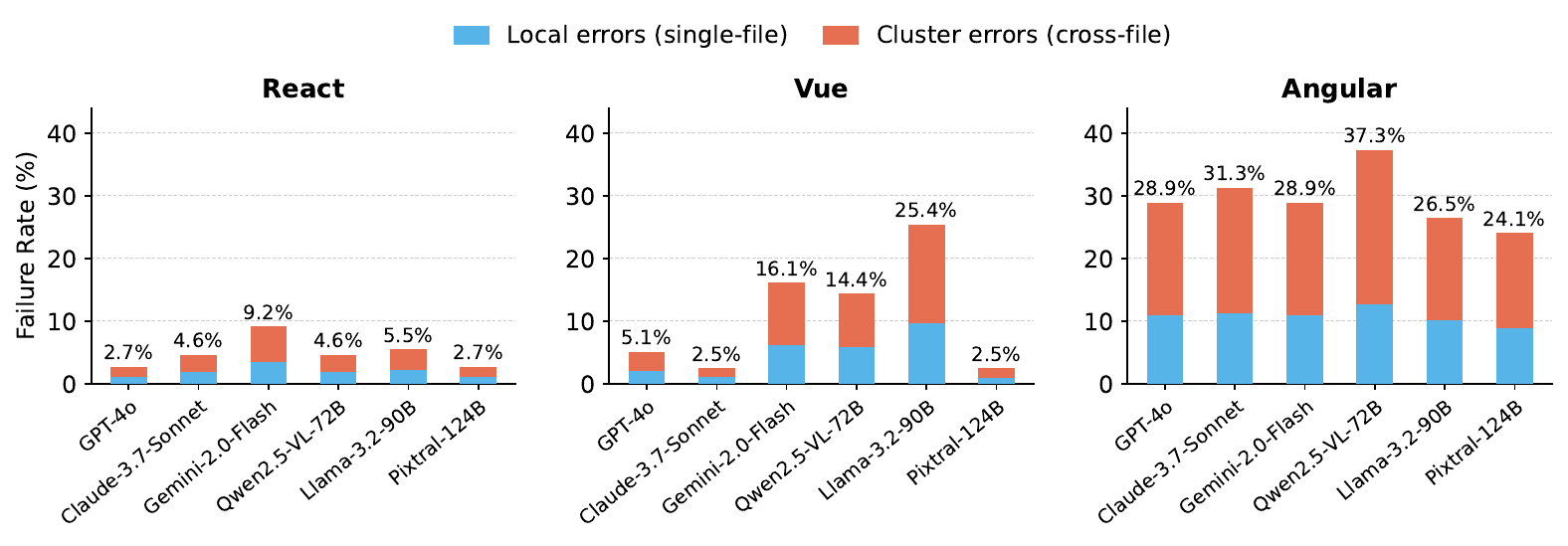}
\vspace{-1.5em}
\caption{Compilation failure rates of six state-of-the-art MLLMs under one-shot generation on React, Vue, and Angular. Failures are decomposed into \textbf{local errors} and \textbf{cluster errors}.}
\vspace{-1em}
\label{fig:error_dist_intro}
\end{figure}

To address these challenges, we study executable multi-framework front-end code generation and propose MESCoder, a tree-structured self-corrective generation framework. Our key idea is to first extract a structural scaffold from the input screenshot, converting large project generation into a sequence of localized subtree expansion and assembly decisions, and then introduce a unified Project Agent that continuously revises its own partial project under runtime sandbox feedback. Rather than treating correction as a post-hoc repair stage, MESCoder integrates self-correction into the generation loop itself, enabling the model to dynamically decide whether to continue construction, perform local repair, or escalate to cluster-level repair when local edits become insufficient. To stably learn this behavior, we further adopt a two-stage training strategy: we first initialize the policy offline to acquire basic multi-framework generation and repair priors, and then refine it with online reinforcement learning to optimize long-horizon executability and final page quality. In this way, we formulate multi-framework front-end generation as a framework-conditioned project state transition process, with self-correction serving as the key mechanism for improving cross-framework executability and the stability of complex front-end generation. 

Our contributions are summarized as follows:
\begin{itemize}[leftmargin=0.5cm]
  \item We propose MESCoder, a tree-structured self-corrective generation framework that decomposes project construction into localized subtree and assembly decisions and equips a unified Project Agent with runtime-feedback-driven self-correction.
  \item We develop a two-stage training strategy that combines offline policy initialization with online reinforcement learning, enabling to optimize long-horizon executability and generation quality.
  \item Extensive experiments demonstrate that MESCoder improves code executability across multiple frameworks while narrowing visual similarity comparable to HTML-based generation.
\end{itemize}

\section{Related Work}

\subsection{WebUI Code Generation}

Generating front-end code from visual inputs has been studied 
extensively. Early methods used classical vision pipelines to 
reverse-engineer UI 
elements~\cite{nguyen2015reverse,moran2018machine}, while 
large-scale datasets such as Rico~\cite{deka2017rico} later enabled 
data-driven approaches. 
Beltramelli~\cite{beltramelli2018pix2code} proposed pix2code, a 
CNN-LSTM model that produces code tokens from GUI screenshots, and 
Chen et al.~\cite{chen2018ui} concurrently introduced a neural 
machine translator for GUI skeletons. 
The emergence of MLLMs has shifted this line of work toward 
fine-tuning on screenshot-code pairs at scale. 
WebSight~\cite{laurenccon2024unlocking} provides two million 
synthetic HTML--screenshot pairs, with parallel efforts such as 
Sketch2Code~\cite{li2025sketch2code} extending inputs to hand-drawn 
sketches. Design2Code~\cite{si2025design2code} introduces the first 
real-world benchmark of 484 webpages with fine-grained visual and 
structural metrics, complemented by IW-Bench~\cite{guo2025iw} with 
DOM-based element and layout accuracy. 
Follow-up work scales training data to over 
one~\cite{yun2024web2code} and 2.56 
million~\cite{gui2025webcode2m} instances, introducing TreeBLEU for 
structural recall, while Interaction2Code~\cite{xiao2025interaction2code} 
extends evaluation to dynamic interactions and 
DesignBench~\cite{xiao2025designbench} to React, Vue, and Angular. 
On the method side, recent approaches improve generation through 
divide-and-conquer decomposition~\cite{wan2025divide}, hierarchical 
generation~\cite{gui2025uicopilot}, modular multi-agent 
pipelines~\cite{jiang2025screencoder}, layout-guided 
prompting~\cite{wu2025mllm}, and structure-aware 
fine-tuning~\cite{liang2025waffle}. 
Closer to our work, MRWeb~\cite{wan2024mrweb} extends generation 
beyond single pages via explicit resource lists, and 
DesignCoder~\cite{chen2025designcoder} couples hierarchical 
generation with self-correction. 
Despite this progress, existing work remains \emph{HTML-centered}, 
targeting self-contained pages that render without compilation. 
Multi-framework generation for React, Vue, and Angular instead 
requires producing multi-file executable projects whose correctness 
depends on whether the project as a whole compiles and 
runs---a setting that prior methods do not address.

\subsection{Self-correction in Code Generation}
Self-correction has emerged as a central capability for code 
generation, allowing models to revise their own outputs based on 
feedback rather than committing to a single one-shot 
prediction~\cite{pan2023automatically}. 
Early prompting-based methods established the basic loop: 
Self-Refine~\cite{madaan2023self} iteratively rewrites a 
program from natural-language critiques produced by the same model, 
Reflexion~\cite{shinn2023reflexion} adds an episodic memory of 
verbal reflections distilled from environment feedback, and 
Self-Debug~\cite{chen2023teaching} grounds the loop in 
execution traces by teaching LLMs rubber-duck-style debugging. 
However, intrinsic self-correction without external signals has 
been shown to plateau or even degrade reasoning 
quality~\cite{huang2023large}, motivating a shift toward 
training-based correction with executable feedback. 
CodeRL~\cite{le2022coderl} and StepCoder~\cite{dou2024stepcoder} 
optimize generation policies against unit-test or compiler 
rewards, while RLEF~\cite{gehring2024rlef} reinforces multi-turn 
LLM agents directly with execution feedback inside the rollout. 
SCoRe~\cite{kumar2024training} introduces a two-stage RL recipe to 
counter behavior collapse and distribution shift in self-correction, 
and \(\mu\)CODE~\cite{jain2025multi} casts multi-turn code generation 
as a single-step decision problem with verifier-guided rewards. 
For small models, CoCoS~\cite{cho2025self} aligns reflective revision 
with parameter-efficient online RL. 
On the structural side, CodeTree~\cite{li2025codetree} performs 
agent-guided tree search over generation and refinement, and 
AlphaCodium~\cite{ridnik2024code} reframes generation as 
a test-driven flow with iterative public-test refinement, while 
DesignCoder~\cite{chen2025designcoder} couples hierarchical UI 
generation with a self-correction loop. 
Despite this progress, existing self-correction methods remain 
\emph{single-file-centric}, treating each program as a 
self-contained unit revised in isolation. 
Multi-framework front-end projects, in contrast, demand correction over a graph of files  whose compilability depends on cross-file imports, bindings, and 
component assembly---a setting where local edits often cannot 
restore project-level coherence and that prior methods do not address.

\section{Methodology}
\label{sec:method}

\begin{figure*}
\centering
\includegraphics[width=0.99\linewidth]{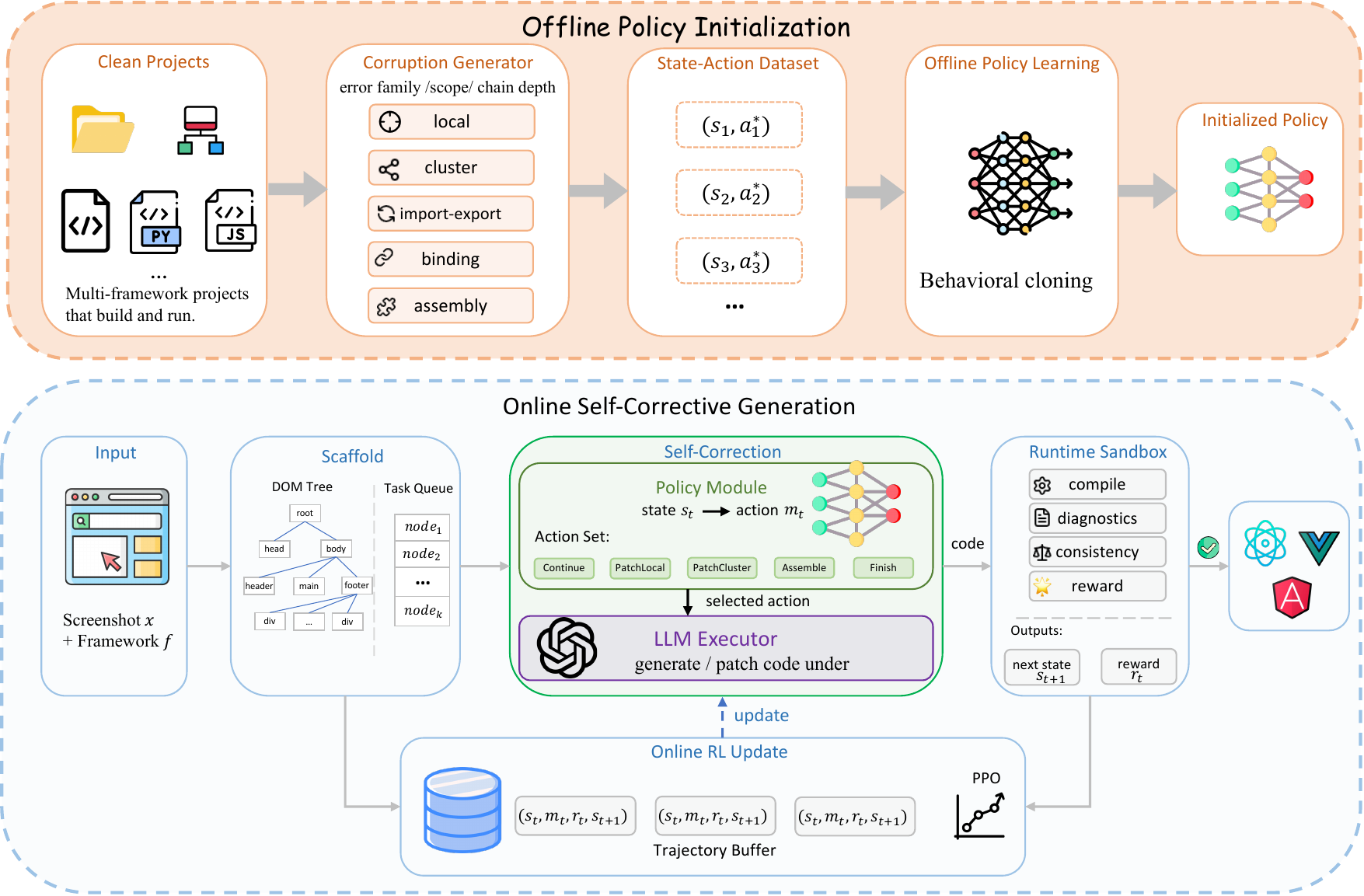}
\caption{The two-stage learning process of MESCoder.}
\vspace{-1em}
\label{fig:mescoder}
\end{figure*}

\subsection{Overview}
\label{sec:method_overview}

Given a webpage screenshot $x$ and a target front-end framework $f \in \{\text{React},\text{Vue},\text{ Angular}\}$, our goal is to generate a complete front-end project $P_T$ that is executable under $f$ and visually consistent with $x$. 
Fig \ref{fig:mescoder} illustrates the learning process of MESCoder, which contains four components: 
A fixed scaffold pre-processor extracts a layout tree and a task queue from the screenshot. 
A policy module selects a high-level action according to the current project state. 
An LLM executor performs the selected action by generating or patching code. 
A runtime sandbox executes the intermediate project, produces diagnostic feedback, and computes executability-oriented rewards.
At step $t$, the interaction is written as
\begin{equation}
m_t \sim \pi_\phi(\cdot \mid s_t), 
\qquad
\Delta_t = \mathrm{LLM}(s_t, m_t),
\qquad
(P_{t+1}, d_{t+1},...) = \mathcal{E}(P_t, \Delta_t, f),
\end{equation}
where $s_t$ is the current state, $m_t$ is the high-level action, $\Delta_t$ is the concrete code edit, $d_{t+1}$ is the diagnostic summary, and $\mathcal{E}$ denotes the framework-aware execution environment.

The learning process has two stages. First, we initialize the policy offline using state-action data. Second, we refine the policy online with reinforcement learning under sandbox feedback. In the main setting, the LLM executor is kept fixed, and both offline initialization and online reinforcement learning optimize only the action  policy and its value estimator.

\subsection{Tree-Structured Self-Corrective Generation}
\label{sec:tree_agent}

Front-end pages are naturally hierarchical. A full page is composed of regions, containers, repeated blocks, and local components, and these units must eventually be assembled into a coherent project. Directly generating the entire project in one pass makes intermediate errors difficult to localize and repair. We therefore first extract a scaffold from the screenshot $\mathcal{S}(x)=\{T,U,Q\}$, where $T$ is a layout tree, $U$ is a content slot table, and $Q=\{\tau_1,\tau_2,\ldots,\tau_N\}$ is a task queue derived from the tree. The scaffold is fixed and not trained. It provides a structural interface that localizes generation and correction decisions.

At interaction step $t$, the current task pointer is $i_t$, and the active task is $q_t=\tau_{i_t}\in Q$.
The agent observes
\begin{equation}
s_t = \{x, f, q_t, P_t, d_t, h_t, b_t\},
\end{equation}
where $P_t$ is the current partial project, $d_t$ is the diagnostic summary returned by the sandbox, $h_t$ stores recent action-feedback history, and $b_t$ is the remaining budget. The history term captures whether recent actions have improved the project, which is important for identifying repeated local failures and deciding whether broader correction is needed.

The policy selects one action from
\begin{equation}
m_t \in 
\{\textsc{Continue},\textsc{PatchLocal},\textsc{PatchCluster},\textsc{Assemble},\textsc{Finish}\}.
\end{equation}
\textsc{Continue} advances the current task when the partial project is sufficiently healthy. \textsc{PatchLocal} invokes a local repair over a small scope. \textsc{PatchCluster} invokes a broader repair over a small dependency or assembly cluster. \textsc{Assemble} integrates completed child units into the current parent node. \textsc{Finish} terminates the episode when the project is complete and executable.
This action set separates forward construction from self-correction. \textsc{Continue} and \textsc{Assemble} construct the project, while \textsc{PatchLocal} and \textsc{PatchCluster} correct intermediate failures. The policy therefore learns not only whether to repair, but also the appropriate scope of repair.

The policy is a lightweight neural controller:
\begin{equation}
\pi_\phi(m_t \mid s_t)=\mathrm{softmax}(g_\phi(z_t)),
\end{equation}
where $z_t$ is a structured encoding of $s_t$ and $g_\phi$ is the actor network. After an action is selected, the LLM executor generates the concrete code edit:
\begin{equation}
\Delta_t = \mathrm{LLM}(s_t,m_t).
\end{equation}
The executor is action-conditioned. Under \textsc{PatchLocal}, it is instructed to modify only a local scope. Under \textsc{PatchCluster}, it may edit several related files or components. Thus, the LLM handles code realization, while the policy handles self-corrective control.

\subsection{Runtime Sandbox and Executability Feedback}
\label{sec:sandbox}

The runtime sandbox defines the execution environment. 
Given the current project $P_t$, an executor edit $\Delta_t$, the selected action $m_t$, and the target framework $f$, the sandbox applies the edit, runs framework-specific checks, and returns the updated project together with compact execution feedback:
\begin{equation}
(P_{t+1}, d_{t+1}, e_t, e_{t+1}, u_t)=\mathcal{E}(P_t,\Delta_t,f). 
\end{equation}
Here, $\mathcal{E}$ only applies and verifies the edit.
The diagnostic summary $d_{t+1}$ is a compact representation of execution feedback. It records the dominant diagnostic family and the affected files or components, and is used as part of the next state rather than as a separate handcrafted reward term.
The quantities $e_t$ and $e_{t+1}$ denote the numbers of sandbox error clusters before and after applying the edit, measured under the same action-relevant verification scope $\Omega_t=\Omega(s_t,m_t)$. The scope depends on the selected action: local repair is verified within the local edit scope, cluster repair within the related file cluster, assembly within the parent-child assembly scope, and \textsc{Finish} on the full project. This avoids penalizing intermediate partial projects for missing future tasks that have not yet been generated.
The binary signal $u_t\in\{0,1\}$ indicates whether the current action successfully commits a new task. For example, $u_t=1$ if a \textsc{Continue} action completes the current subtree, or if an \textsc{Assemble} action successfully connects a completed child unit to its parent. Otherwise, $u_t=0$.
Since intermediate partial projects may not be fully renderable, visual quality is not used as a dense step-level signal. We compute a terminal quality score $Q_T$ only after a final renderable project is produced.

\subsection{Two-Stage Policy Learning}
\label{sec:policy_learning}

\paragraph{Offline policy initialization.}
The offline stage initializes the action policy before online reinforcement learning. Its goal is not to train the LLM executor or to teach code generation. Instead, it provides a reasonable action prior for the policy module $\pi_\phi(m \mid s)$.

For each clean executable project $P^\star$, we construct normal project-construction states by replaying the tree-guided construction process. These states provide supervision for non-repair actions. If the current node is unfinished and no critical diagnostic is present, the label is \textsc{Continue}. If a subtree is completed and should be integrated into its parent, the label is \textsc{Assemble}. If the root project is complete and strict checks pass, the label is \textsc{Finish}. Thus, clean replay provides examples for the forward construction behavior.
To initialize repair decisions, we generate corrupted states from clean projects. Given $P^\star$, an automatic corruption operator produces
\begin{equation}
\tilde P = \mathcal{C}(P^\star).
\end{equation}
For a corrupted state $s_t$, we define a candidate repair set $\mathcal{M}(s_t)$, typically containing \textsc{PatchLocal} and \textsc{PatchCluster}. For each candidate action $m\in\mathcal{M}(s_t)$, the LLM executor is called $K$ times to generate candidate edits:
\begin{equation}
\Delta_{m,k} = \mathrm{LLM}(s_t,m), \qquad k=1,\ldots,K,
\end{equation}
where $K$ is the number of executor samples used only for offline outcome estimation; it reduces labeling noise caused by stochastic LLM edits.
Each candidate edit is evaluated by the sandbox. Let
\begin{equation}
\hat R(s_t,m)=\max_{k} R(s_t,m,\Delta_{m,k})
\end{equation}
be the best outcome score obtained under action $m$. The offline repair label is then
\begin{equation}
m_t^\star = \arg\max_{m\in\mathcal{M}(s_t)} \hat R(s_t,m).
\end{equation}
This result-driven labeling procedure avoids reducing offline learning to a predefined error-action rule. It assigns action labels according to observed outcomes, so the same diagnostic family may still lead to different repair scopes under different contexts.
The policy is initialized by minimizing
\begin{equation}
\mathcal{L}_{\mathrm{off}}
=
-\sum_{(s_t,m_t^\star)\in\mathcal{D}_{\mathrm{off}}}
\log \pi_\phi(m_t^\star\mid s_t).
\end{equation}
The resulting policy $\pi_{\phi_0}$ provides an action prior for subsequent online learning.

\paragraph{Online reinforcement learning.}
Offline initialization provides a policy prior, but executable project generation remains a long-horizon decision problem. During real generation, actions interact with LLM-generated edits, framework-specific constraints, and accumulated project states. Online reinforcement learning further optimizes the policy under actual sandbox feedback.

Starting from $\pi_{\phi_0}$, the agent interacts with the environment and collects trajectories $(s_t,m_t,r_t,s_{t+1}).$
The online step reward is constructed from the sandbox feedback:
\begin{equation}
r_t^{\mathrm{step}} = \alpha(e_t-e_{t+1}) + \beta u_t - \gamma c_t.
\end{equation}
The first term rewards error reduction in the action-relevant verification scope. The second term rewards successful task commitment. The third term penalizes expensive or invalid actions. The cost $c_t$ depends on the selected action, edit size, verification scope size, and whether the edit can be validly applied.
At the end of an episode, the agent receives
\begin{equation}
R_{\mathrm{final}}=
\begin{cases}
r_{\mathrm{pass}}+\mu Q_T, & \text{SUCCESS} ,\\
r_{\mathrm{partial}}, & m_T= \text{FINISH and ERROR},\\
0, & \text{otherwise}.
\end{cases}
\end{equation}
The complete optimization objective is
\begin{equation}
J(\phi)=
\mathbb{E}_{\tau\sim\pi_\phi}
\left[
\sum_{t=0}^{T-1} r_t^{\mathrm{step}} + R_{\mathrm{final}}
\right].
\end{equation}

We optimize the policy using an actor-critic formulation. The actor is $\pi_\phi(m\mid s)$, while the critic estimates the state value $V_\psi(s)$. In our implementation, we use PPO-style clipped policy optimization for stability. The actor and critic parameters, $\phi$ and $\psi$, are updated during online learning, while the LLM executor remains fixed in the main setting. This keeps reinforcement learning focused on high-level self-corrective decision making rather than token-level code generation.

\section{Experiment}

\subsection{Experimental Setup}
\label{sec:setup}

\paragraph{Benchmarks and Metrics.}
We evaluate on \textbf{DesignBench}~\cite{xiao2025designbench}, the benchmark pairing real-world webpage screenshots with ground-truth executable multi-framework projects. Its \emph{design generation} split contains 310 compilation-verified pages across React, Vue, and Angular. We stratify these by framework and complexity, holding out 200 pages as the test set and reserving the remaining 110 for training. We report CSR (Compilation Success Rate), the fraction of projects passing strict compilation and runtime verification, and CLIP~\cite{si2025design2code}, the visual similarity to the reference image.

\paragraph{Training Data.}
Our training corpus combines the 110 non-test DesignBench pages (compilation-verified, multi-framework) with 2{,}000 projects sampled from \textbf{WebCode2M}~\cite{gui2025webcode2m} (filtered, decomposed into multi-file hierarchies, and deduplicated against the test set). All clean projects are processed by a corruption pipeline $\mathcal{C}$ (Appendix~\ref{app:sandbox_data}) that injects realistic LLM error modes across varying repair scopes, yielding a comprehensive offline corpus of 27{,}430 state-action pairs.

\paragraph{Baselines.}
We compare MESCoder against four baselines that span from  direct prompting to heuristic and learned self-correction:
(1) \textbf{One-Shot}: Generates the full multi-file project in a single LLM forward pass without structural decomposition or post-hoc repair, serving as the vanilla generation reference.
(2) \textbf{Rule Repair}: Augments a scaffold-based generation pipeline with hard-coded, heuristic fix templates for the 20 most frequent compilation errors. This represents state-of-the-art non-learned execution repair.
(3) \textbf{RLEF} (RL from Execution Feedback): Keeps LLM executor frozen and trains an external controller using feedback. The controller modulates the executor through generic generation instructions, but does not use our action space. It serves as a monolithic RL baseline for testing whether sandbox-feedback optimization alone can match MESCoder.
(4) \textbf{Offline-Only}: An ablation baseline that uses our exact scaffold representation and MLP policy trained exclusively on the offline demonstration corpus, omitting the online PPO exploration stage.

\begin{table}[t!]
\centering
\caption{Main results on DesignBench. 
\textit{HTML Ref.} reports the CLIP of the same executor generating a single HTML file, serving as a compilation-free visual reference.
Best multi-framework result per executor in \textbf{bold}. All metrics are averaged over three independent runs.}
\label{tab:main}
\small
\begin{tabular}{llcccccc|c}
\toprule
& & \multicolumn{2}{c}{\textbf{React}} & \multicolumn{2}{c}{\textbf{Vue}} & \multicolumn{2}{c|}{\textbf{Angular}} & \textbf{HTML} \\
\cmidrule(lr){3-4}\cmidrule(lr){5-6}\cmidrule(lr){7-8}\cmidrule(lr){9-9}
\textbf{Executor} & \textbf{Method} & CSR & CLIP & CSR & CLIP & CSR & CLIP & CLIP \\
\midrule
\multirow{3}{*}{GPT-4o}
 & \textit{HTML Ref.} & -- & -- & -- & -- & -- & -- & \textit{0.784} \\
 & One-Shot & 97.3 & 0.758 & 94.9 & 0.769 & 71.1 & 0.591 & -- \\
 & \textbf{MESCoder} & \textbf{99.1} & \textbf{0.766} & \textbf{97.8} & \textbf{0.779} & \textbf{89.2} & \textbf{0.661} & -- \\
\midrule
\multirow{3}{*}{Claude-3.7-Sonnet}
 & \textit{HTML Ref.} & -- & -- & -- & -- & -- & -- & \textit{0.853} \\
 & One-Shot & 95.4 & 0.813 & 97.5 & 0.827 & 68.7 & 0.597 & -- \\
 & \textbf{MESCoder} & \textbf{98.3} & \textbf{0.825} & \textbf{99.1} & \textbf{0.832} & \textbf{87.1} & \textbf{0.674} & -- \\
\midrule
\multirow{3}{*}{Gemini-2.0-Flash}
 & \textit{HTML Ref.} & -- & -- & -- & -- & -- & -- & \textit{0.749} \\
 & One-Shot & 90.8 & 0.754 & 83.9 & 0.694 & 71.1 & 0.606 & -- \\
 & \textbf{MESCoder} & \textbf{96.1} & \textbf{0.777} & \textbf{92.4} & \textbf{0.722} & \textbf{87.3} & \textbf{0.663} & -- \\
\midrule
\multirow{3}{*}{Qwen2.5-VL-72B}
 & \textit{HTML Ref.} & -- & -- & -- & -- & -- & -- & \textit{0.809} \\
 & One-Shot & 95.4 & 0.783 & 85.6 & 0.681 & 62.7 & 0.519 & -- \\
 & \textbf{MESCoder} & \textbf{97.6} & \textbf{0.792} & \textbf{93.1} & \textbf{0.706} & \textbf{82.4} & \textbf{0.572} & -- \\
\midrule
\multirow{3}{*}{Llama-3.2-90B}
 & \textit{HTML Ref.} & -- & -- & -- & -- & -- & -- & \textit{0.724} \\
 & One-Shot & 94.5 & 0.698 & 74.6 & 0.537 & 73.5 & 0.528 & -- \\
 & \textbf{MESCoder} & \textbf{97.2} & \textbf{0.710} & \textbf{86.8} & \textbf{0.577} & \textbf{88.1} & \textbf{0.573} & -- \\
\midrule
\multirow{3}{*}{Pixtral-124B}
 & \textit{HTML Ref.} & -- & -- & -- & -- & -- & -- & \textit{0.796} \\
 & One-Shot & 97.3 & 0.744 & 97.5 & 0.776 & 75.9 & 0.629 & -- \\
 & \textbf{MESCoder} & \textbf{98.8} & \textbf{0.750} & \textbf{99.0} & \textbf{0.781} & \textbf{91.1} & \textbf{0.694} & -- \\
\bottomrule
\end{tabular}
\vspace{-1em}
\end{table}

\subsection{Main Results}
\label{sec:main_results}

Table~\ref{tab:main} reports end-to-end results across three frameworks and six LLM executors. MESCoder consistently raises CSR over every executor--framework combination. For React and Vue, where One-Shot baselines are already strong, the gains are moderate yet consistent across all executors. For Angular, where strict TypeScript module-registration constraints make single-pass generation most brittle, the improvements are markedly larger. Crucially, even the strongest executors benefit substantially, demonstrating that cluster-scope repair addresses failure modes that raw executor capability alone cannot resolve. The same policy architecture and training recipe are applied identically across all six executors, so the uniform pattern of improvement constitutes strong evidence that MESCoder's benefit is architectural rather than executor-specific. Alongside the CSR improvements, CLIP rises in proportion: Angular sees the largest visual-fidelity gains, consistent with its steeper executability improvement, while React and Vue show smaller but reliable increases, confirming that executability and visual quality improve together rather than trading off. The \textit{HTML Ref.}\ column provides an executor-matched upper bound; MESCoder narrows the gap to this ceiling substantially on Angular while approaching it on React and Vue. A complexity-stratified breakdown (Appendix~\ref{app:difficulty_detail}, Table~\ref{tab:difficulty}) further confirms that MESCoder's advantage is most pronounced on Hard samples, where cluster-level correction becomes critical when failure chains span multiple files.

\subsection{Policy Learning and Behavior}
\label{sec:policy_behavior}

We analyze how the two-stage training procedure shapes the policy's self-correction behavior across training dynamics and scope escalation.

\begin{figure}[t]
\centering
\begin{minipage}{0.48\linewidth}
\centering
\includegraphics[width=\linewidth]{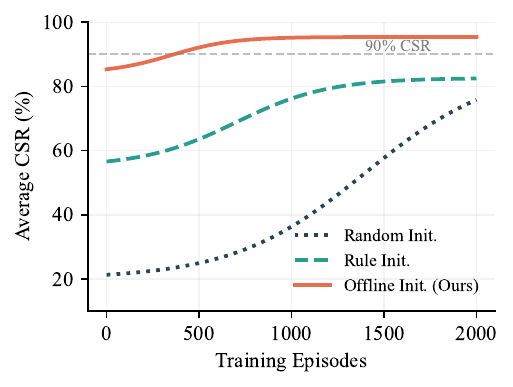}
\end{minipage}
\hfill
\begin{minipage}{0.48\linewidth}
\centering
\includegraphics[width=\linewidth]{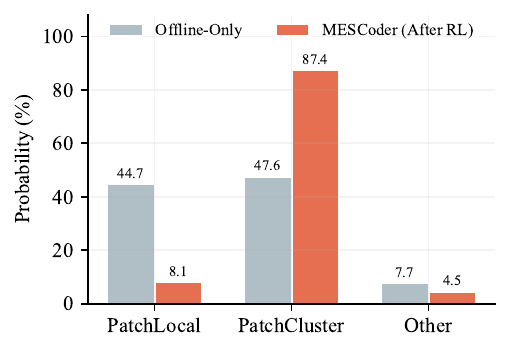}
\end{minipage}
\caption{\textbf{Left}: Average CSR on the validation set during the online RL stage under three initialization conditions: behavior-cloning (offline) initialization, hand-coded rule initialization (rule policy used as a parameter seed, then fine-tuned with PPO), and random initialization. Offline initialization converges faster and plateaus higher, confirming the value of the offline prior. \textbf{Right}: Action distribution on stagnated-cluster states before and after RL fine-tuning. RL substantially increases the \textsc{PatchCluster} probability while reducing spurious \textsc{PatchLocal} decisions on these hard states.}
\vspace{-1em}
\label{fig:policy_training}
\end{figure}

\paragraph{Training dynamics and action shift.}
Figure~\ref{fig:policy_training} illustrates training dynamics and the resulting action distribution shift. The offline-initialized policy reaches a stable, high-CSR plateau substantially faster than both random and rule-initialized baselines, confirming that the behavior-cloning stage provides a meaningful action prior that makes online RL tractable. The rule-initialized agent converges more slowly and stabilizes at a lower final CSR, as rigid heuristic rules cannot capture context-dependent scope decisions. On \emph{stagnated-cluster} states---where repeated local repair has yielded no error reduction---the offline-only policy distributes probability nearly equally between \textsc{PatchLocal} and \textsc{PatchCluster}; after online RL, \textsc{PatchCluster} rises to a clear dominant position while \textsc{PatchLocal} drops sharply. Crucially, the shift is selective:  
this suggests that RL primarily increases cluster repair on states where local repair has stagnated,
confirming that RL has learned to escalate scope \emph{when and only when} local repair has demonstrably failed.

\paragraph{Scope escalation analysis.}
To directly quantify escalation behavior, we curate a diagnostic subset of stagnated-cluster episodes and compare methods on escalation rate, final CSR, average steps, and invalid patch count (Appendix~\ref{app:escalation_detail}), together with two oracle variants: \emph{Always-Local} (never escalates) and \emph{Always-Cluster} (escalates unconditionally). \emph{Always-Local} achieves the lowest CSR, unable to escape local stagnation; \emph{Always-Cluster} achieves a perfect escalation rate but lower CSR than MESCoder, since indiscriminate broad edits disturb previously healthy files. MESCoder achieves the highest escalation rate among learned methods and the best final CSR, demonstrating that online RL teaches selective escalation.

\subsection{Error-Dimension Analysis}
\label{sec:error_analysis}

\begin{figure}[h]
\centering
\begin{minipage}{0.48\linewidth}
\centering
\includegraphics[width=\linewidth]{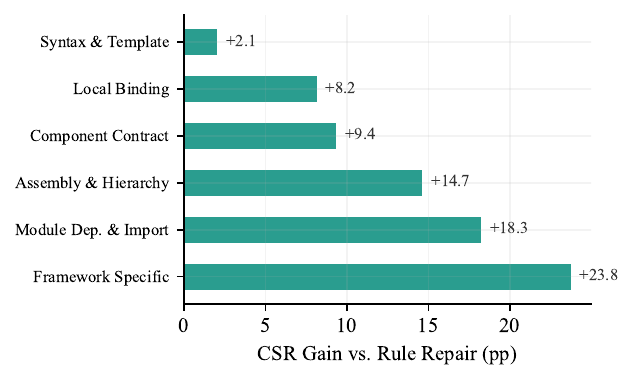}
\end{minipage}
\hfill
\begin{minipage}{0.48\linewidth}
\centering
\includegraphics[width=\linewidth]{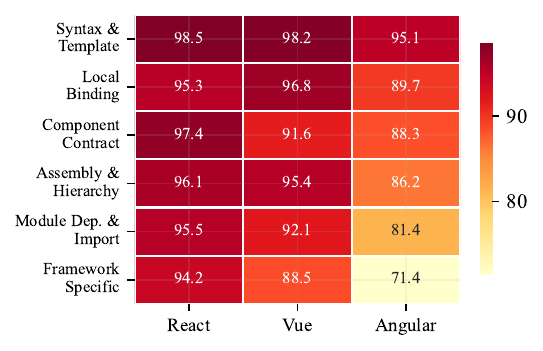}
\end{minipage}
\caption{\textbf{Left}: CSR gain over Rule Repair per error family (GPT-4o, injected-error subset). \textbf{Right}: Per-cell CSR across framework--error combinations. Angular shows the steepest difficulty gradient.}
\label{fig:error_analysis}
\vspace{-0.5em}
\end{figure}

Figure~\ref{fig:error_analysis} characterizes MESCoder's gains across error families and frameworks. The left panel shows CSR gain over Rule Repair per error family (GPT-4o, injected-error subset): gains increase from syntactic failures (\emph{Syntax \& Template}) through intra-component errors (\emph{Local Binding}, \emph{Component Contract}) to cross-file coordination failures (\emph{Assembly \& Hierarchy}, \emph{Module Dep. \& Import/Export}, \emph{Framework-Specific Engineering})---validating that \textsc{PatchCluster} addresses the failure modes single-file repair cannot resolve. The right panel shows per-cell CSR across framework--error-family combinations; Angular exhibits the steepest difficulty gradient.

\subsection{Ablation Study}
\label{sec:ablation}

Table~\ref{tab:ablation} presents a systematic ablation of MESCoder's key design choices, using GPT-4o as the executor and reporting average CSR across the three frameworks.

\begin{table}[h]
\centering
\caption{Ablation study on average CSR (\%) across React, Vue, and Angular (GPT-4o executor). Each row removes or replaces one design component from the full MESCoder system.}
\label{tab:ablation}
\renewcommand{\arraystretch}{0.75}
\begin{tabular}{lcc}
\toprule
\textbf{Variant} & Avg CSR & $\Delta$ vs. Full \\
\midrule
Full MESCoder                              & 95.4 & --   \\
\midrule
\multicolumn{3}{l}{\textit{Architecture ablations}} \\
\quad w/o scaffold (flat generation)       & 80.8 & $-$14.6 \\
\quad replace MLP policy w/ rule policy    & 82.6 & $-$12.8 \\
\quad w/o \textsc{PatchCluster} action     & 85.5 & $-$9.9  \\
\quad w/o \textsc{Assemble} action         & 89.7 & $-$5.7  \\
\quad w/o history features $h_t$           & 91.5 & $-$3.9  \\
\midrule
\multicolumn{3}{l}{\textit{Training ablations}} \\
\quad w/o offline init.\ (online RL only)  & 86.2 & $-$9.2  \\
\quad w/o online RL (= Offline-Only)       & 84.1 & $-$11.3 \\
\quad w/o action-relevant scope $\Omega_t$ & 88.3 & $-$7.1  \\
\quad w/o PPO clipping                     & 92.2 & $-$3.2  \\
\quad w/o entropy regularization           & 92.7 & $-$2.7  \\
\midrule
\multicolumn{3}{l}{\textit{Reward ablations}} \\
\quad w/o error-reduction reward           & 81.2 & $-$14.2 \\
\quad w/o task-commitment reward $u_t$     & 91.6 & $-$3.8  \\
\quad w/o terminal quality $Q_T$           & 93.5 & $-$1.9  \\
\bottomrule
\end{tabular}
\vspace{-1em}
\end{table}

\paragraph{Architecture and action space.}
Removing the scaffold causes the largest CSR drop, confirming that decomposing project generation into localized subtree decisions is fundamental; replacing the MLP policy with a rule-based selector causes a comparably large drop, showing that context-dependent scope decisions cannot be captured by fixed heuristics. Removing \textsc{PatchCluster} leads to a substantially larger degradation than removing \textsc{Assemble}, validating that the ability to escalate to cross-file cluster repair is the more critical capability once local repair stagnates. Removing $h_t$ also incurs a penalty, confirming history-aware encoding is necessary for reliable stagnation detection.

\paragraph{Training, scope, and reward.}
Both removing online RL and removing offline initialization cause substantial CSR drops, with online RL contributing slightly more, reflecting complementary failure modes. Replacing $\Omega_t$ with full-project verification destabilizes training. Among rewards, $\alpha(e_t - e_{t+1})$ is dominant; $\beta u_t$ provides secondary support; $Q_T$ contributes a smaller but non-negligible gain, primarily by encouraging generation quality at episode termination.
Inference efficiency (Appendix~\ref{app:efficiency_detail}, Table~\ref{tab:efficiency}) and qualitative case studies (Appendix~\ref{app:qualitative_detail}) are provided in the appendix.

\section{Conclusion}

We presented MESCoder, a tree-structured self-corrective generation framework that addresses the challenge of executable multi-framework front-end code synthesis. By decomposing project construction into scaffold-guided subtree expansions and equipping a unified Project Agent with a learned policy for context-sensitive local and cluster-level repair under sandbox execution feedback, MESCoder transforms self-correction from a post-hoc step into an integral part of the generation loop. Experiments across React, Vue, and Angular with six LLM executors demonstrate consistent CSR improvements, most pronounced on Angular, where cross-file dependency failures are prevalent, while visual fidelity is preserved throughout.

\bibliographystyle{plainnat}
\bibliography{reference}
\clearpage
\appendix

\section{Difficulty-Stratified Results}
\label{app:difficulty_detail}

\paragraph{Complexity stratification protocol.}
We partition the 200 test pages into three difficulty tiers using scaffold tree depth $D$ and inter-file dependency edge count $E$.
A page is \textbf{Easy} if $D \leq 3$ and $E \leq 4$; \textbf{Hard} if $D \geq 6$ or $E \geq 10$; and \textbf{Medium} otherwise.
This yields 68 Easy, 87 Medium, and 45 Hard.

\paragraph{Aggregate difficulty results.}
Table~\ref{tab:difficulty} reports CSR and CLIP averaged across all three frameworks.
MESCoder's CSR advantage grows monotonically from Easy through Medium to Hard, directly reflecting the increasing importance of cluster-level repair as cross-file dependencies deepen.
On Hard pages, Offline-Only drops sharply because its policy---trained without online exploration---consistently misclassifies stagnated cluster failures as local ones and exhausts its step budget.

\begin{table}[h]
\centering
\caption{Results stratified by page complexity (GPT-4o executor, averaged across React / Vue / Angular). Complexity is defined by scaffold tree depth $D$ and inter-file dependency edge count $E$.}
\label{tab:difficulty}
\begin{tabular}{lcccccc}
\toprule
\textbf{Method} & Easy CSR & Med CSR & Hard CSR & Easy CLIP & Med CLIP & Hard CLIP \\
\midrule
One-Shot      & 95.1 & 87.4 & 77.8 & 0.712 & 0.669 & 0.583 \\
Rule Repair   & 96.3 & 90.1 & 80.3 & 0.726 & 0.683 & 0.601 \\
RLEF          & 96.8 & 92.6 & 85.2 & \textbf{0.741} & 0.697 & 0.629 \\
Offline-Only  & 96.6 & 81.6 & 70.0 & 0.737 & 0.674 & 0.598 \\
\textbf{MESCoder}  & \textbf{98.6} & \textbf{95.3} & \textbf{91.7} & 0.739 & \textbf{0.701} & \textbf{0.632} \\
\bottomrule
\end{tabular}
\end{table}

\paragraph{Per-framework difficulty breakdown.}
Table~\ref{tab:difficulty_per_fw} disaggregates Hard-tier results by framework.
Angular shows the steepest difficulty gradient across all methods, reflecting that its strict module-wiring constraints make deep-hierarchy pages disproportionately failure-prone; React and Vue Hard pages are substantially easier, with all methods achieving higher CSR.

\begin{table}[h]
\centering
\caption{Hard-tier CSR (\%) per framework (GPT-4o executor).}
\label{tab:difficulty_per_fw}
\begin{tabular}{lccc}
\toprule
\textbf{Method} & Hard React & Hard Vue & Hard Angular \\
\midrule
One-Shot     & 88.6 & 86.3 & 58.4 \\
Rule Repair  & 90.2 & 88.7 & 62.1 \\
RLEF         & 93.4 & 91.8 & 70.3 \\
Offline-Only & 82.6 & 80.1 & 47.2 \\
\textbf{MESCoder} & \textbf{96.3} & \textbf{95.1} & \textbf{83.7} \\
\bottomrule
\end{tabular}
\end{table}

\section{Offline Data Construction}
\label{app:sandbox_data}

\paragraph{Clean project sources.}
We use 110 DesignBench training pages plus 2{,}000 pages from WebCode2M.
The WebCode2M pages are filtered for compilation correctness, decomposed into multi-file framework hierarchies, and deduplicated against the test set.

\paragraph{Corruption operators.}
The corruption pipeline $\mathcal{C}$ applies six operators, each targeting a distinct error family, to generate realistic error-inducing states from clean projects (Table~\ref{tab:corruption_ops}):

\begin{table}[h]
\centering
\caption{Corruption operators used in offline data construction.}
\label{tab:corruption_ops}
\begin{tabular}{llp{7.5cm}}
\toprule
\textbf{Operator} & \textbf{Target Family} & \textbf{Description} \\
\midrule
$\mathcal{C}_\text{syn}$  & Syntax \& Template      & Removes closing tags, introduces invalid JSX attribute syntax, or misbalances template braces. \\
$\mathcal{C}_\text{bind}$ & Local Binding           & Renames a prop or variable reference to an undefined name within a single component. \\
$\mathcal{C}_\text{cont}$ & Component Contract      & Changes a component's exported interface without updating its parent call site. \\
$\mathcal{C}_\text{asm}$  & Assembly \& Hierarchy   & Removes or misroutes an import connecting a child component to its parent. \\
$\mathcal{C}_\text{mod}$  & Module Dep. \& Import       & Renames or removes an export without updating downstream import sites, producing missing-export or name-mismatch errors. \\
$\mathcal{C}_\text{fw}$   & Framework Specific      & Removes an Angular component from \texttt{@NgModule}, or introduces a Vue \texttt{setup()} hook incompatible with Options API. \\
\bottomrule
\end{tabular}
\end{table}

\paragraph{Result-driven repair labeling.}
For each corrupted state, the repair label is assigned by calling the LLM executor $K{=}4$ times per candidate action and selecting the action that achieves the highest sandbox repair score.
After filtering, we retain 27{,}430 labeled pairs.

\section{Implementation Details}
\label{sec:impl_details}

The scaffold preprocessor uses a pretrained visual detector and rule-based grouping to construct hierarchical layout trees. The lightweight MLP policy encodes the multimodal state $s_t$ via frozen encoders into a latent vector, then outputs an action distribution over the hierarchical action space. We evaluate across six frozen LLM executors: GPT-4o, Claude-3.7-Sonnet, Gemini-2.0-Flash, Qwen2.5-VL-72B, Llama-3.2-90B, and Pixtral-124B. Experiments are run on 8$\times$A100 GPUs, and all results are averaged over three independent runs.

\section{Scope Escalation Detail}
\label{app:escalation_detail}

Table~\ref{tab:escalation} compares five strategies on escalation rate, final CSR, average steps, and invalid patch count.
The two oracle variants bound the range: \emph{Always-Local} stagnates without ever escalating; \emph{Always-Cluster} escalates unconditionally but disturbs healthy files, yielding lower CSR than MESCoder.
Among learned methods, MESCoder achieves the highest escalation rate and the best CSR, demonstrating that online RL teaches context-sensitive scope escalation.

\begin{table}[h]
\centering
\caption{Scope escalation analysis on the stagnated-cluster diagnostic subset.}
\label{tab:escalation}
\begin{tabular}{lcccc}
\toprule
\textbf{Method} & Escal.\ Rate (\%) & CSR (\%) & Avg Steps & Inv.\ Patches \\
\midrule
\textit{Always-Local}   & 0.0   & 21.7 & 15.2 & 6.12 \\
\textit{Always-Cluster} & 100.0 & 44.8 & 9.3  & 1.89 \\
\midrule
Rule Repair             & 26.3  & 39.2 & 12.4 & 3.71 \\
Offline-Only            & 57.8  & 58.3 & 10.8 & 2.43 \\
\textbf{MESCoder}       & \textbf{90.3} & \textbf{76.2} & \textbf{11.3} & \textbf{1.12} \\
\bottomrule
\end{tabular}
\end{table}

\paragraph{Per-framework escalation breakdown.}
Table~\ref{tab:escalation_per_fw} disaggregates escalation rate and CSR by framework for MESCoder and the two strongest baselines.
Angular exhibits the highest stagnated-cluster rate (52 of the 87 episodes come from Angular), consistent with its strict module requirements.
MESCoder's escalation rate on Angular stagnated-cluster episodes (91.4\%) is the highest across all frameworks, while its CSR (69.8\%) reflects that even correct escalation does not fully resolve all Angular failures---the deepest dependency chains remain beyond the episode horizon.
React and Vue show lower stagnation frequencies (18 and 17 episodes respectively) and higher post-escalation CSR, as their dependency graphs are shallower and cluster edits more reliably sufficient.

\begin{table}[h]
\centering
\caption{Per-framework escalation rate and CSR on the stagnated-cluster subset (GPT-4o executor). Episode counts reflect the frequency of stagnated-cluster episodes per framework.}
\label{tab:escalation_per_fw}
\begin{tabular}{llccc}
\toprule
\textbf{Method} & \textbf{Metric} & React ($n{=}18$) & Vue ($n{=}17$) & Angular ($n{=}52$) \\
\midrule
\multirow{2}{*}{Rule Repair}
  & Escal.\ Rate (\%) & 27.8 & 29.4 & 25.0 \\
  & CSR (\%)          & 44.4 & 47.1 & 34.6 \\
\midrule
\multirow{2}{*}{Offline-Only}
  & Escal.\ Rate (\%) & 66.7 & 64.7 & 52.3 \\
  & CSR (\%)          & 72.2 & 70.6 & 49.4 \\
\midrule
\multirow{2}{*}{\textbf{MESCoder}}
  & Escal.\ Rate (\%) & \textbf{88.9} & \textbf{88.2} & \textbf{91.4} \\
  & CSR (\%)          & \textbf{88.9} & \textbf{82.4} & \textbf{69.8} \\
\bottomrule
\end{tabular}
\end{table}

\section{Inference Efficiency Detail}
\label{app:efficiency_detail}

Table~\ref{tab:efficiency} profiles the inference cost of each method (GPT-4o, averaged across frameworks).
MESCoder's sandbox invocations exceed its step count because each \textsc{PatchCluster} action triggers pre- and post-patch verification.
RLEF and Offline-Only operate at comparable interaction depth yet achieve notably lower CSR, confirming that structured two-stage training is the key differentiator.

\begin{table}[h]
\centering
\caption{Inference efficiency on DesignBench averaged across React / Vue / Angular.}
\label{tab:efficiency}
\begin{tabular}{lccccc}
\toprule
\textbf{Method} & Avg CSR & Avg Steps & Sandbox Calls & LLM Calls & Wall Time (s) \\
\midrule
One-Shot      & 87.8 & 1.0  & 1.0   & 1.0  & 18.3  \\
Rule Repair   & 89.4 & 4.1  & 8.6   & 4.1  & 52.7  \\
RLEF          & 91.8 & 7.3  & 7.3   & 7.3  & 94.2  \\
Offline-Only  & 84.1 & 8.4  & 8.4   & 8.4  & 109.1 \\
\textbf{MESCoder} & \textbf{95.4} & \textbf{10.2} & \textbf{14.7} & \textbf{10.2} & \textbf{134.6} \\
\bottomrule
\end{tabular}
\end{table}

\section{Qualitative Analysis}
\label{app:qualitative_detail}

\paragraph{Error injection and repair.}
We present three concrete case studies drawn from DesignBench to illustrate how operator $\mathcal{C}$ injects representative errors and how MESCoder resolves each.

\begin{center}
\textit{Listing 1. Wrong React code ($\mathcal{C}_\text{bind}$).}
\end{center}
\vspace{-0.5em}
\begin{lstlisting}[numbers=left, xleftmargin=1.5em,
  basicstyle=\small\ttfamily, breaklines=true, frame=single,
  label={lst:react_wrong},
  language=Java,
  morekeywords={import,from,export,const,function,return,useState},
  commentstyle=\color{gray},
  keywordstyle=\color{blue},
  stringstyle=\color{orange!70!black}]
import { useState } from "react";

export default function App() {
  const [todos, setTodos] = useState([
    { text: "Write a new blog post.", completed: false },
    { text: "Pick up laundry.",       completed: false },
  ]);

  const handleToggleComplete = (index) => {
    const updated = todos.map((todo, i) =>
      i === index
        ? { ...todo, complete: !todo.complete }  // renamed by C_bind
        : todo
    );
    setTodos(updated);
  };

  return (
    <ul>
      {todos.map((todo, i) => (
        <li key={i} onClick={() => handleToggleComplete(i)}
          style={{ textDecoration: todo.completed ? "line-through" : "none" }}>
          {todo.text}
        </li>
      ))}
    </ul>
  );
}
\end{lstlisting}

\newpage
\begin{center}
\textit{Listing 2. Fixed React code.}
\end{center}
\vspace{-0.5em}
\begin{lstlisting}[numbers=left, xleftmargin=1.5em,
  basicstyle=\small\ttfamily, breaklines=true, frame=single,
  label={lst:react_fixed},
  language=Java,
  morekeywords={import,from,export,const,function,return,useState},
  commentstyle=\color{gray},
  keywordstyle=\color{blue},
  stringstyle=\color{orange!70!black}]
import { useState } from "react";

export default function App() {
  const [todos, setTodos] = useState([
    { text: "Write a new blog post.", completed: false },
    { text: "Pick up laundry.",       completed: false },
  ]);

  const handleToggleComplete = (index) => {
    const updated = todos.map((todo, i) =>
      i === index
        ? { ...todo, completed: !todo.completed }  // Fixed
        : todo
    );
    setTodos(updated);
  };

  return (
    <ul>
      {todos.map((todo, i) => (
        <li key={i} onClick={() => handleToggleComplete(i)}
          style={{ textDecoration: todo.completed ? "line-through" : "none" }}>
          {todo.text}
        </li>
      ))}
    </ul>
  );
}
\end{lstlisting}

\newpage
\begin{center}
\textit{Listing 3. Wrong Vue code ($\mathcal{C}_\text{cont}$).}
\end{center}
\vspace{-0.5em}
\begin{lstlisting}[numbers=left, xleftmargin=1.5em,
  basicstyle=\small\ttfamily, breaklines=true, frame=single,
  label={lst:vue_wrong},
  language=Java,
  morekeywords={import,from,export,const,defineProps},
  commentstyle=\color{gray},
  keywordstyle=\color{blue},
  stringstyle=\color{orange!70!black}]
const ResourceCard = (props: {
  iconName: string;  // renamed by C_cont
  title: string;
  description: string;
}) => (
  <div class="bg-white shadow rounded-lg p-6">
    <i class={`fas fa-${props.iconName} text-blue-500`}></i>  // renamed by C_cont
    <h2 class="text-lg font-medium">{props.title}</h2>
    <p class="text-sm text-gray-500">{props.description}</p>
  </div>
);

<div class="grid grid-cols-1 md:grid-cols-3 gap-6">
  <ResourceCard icon="file-text"
                title="Blog"
                description="Explore our learnings..." />
  <ResourceCard icon="clipboard-check"
                title="Changelog"
                description="Stay updated..." />
  <ResourceCard icon="users"
                title="Community"
                description="Join the discussion..." />
</div>
\end{lstlisting}

\begin{center}
\textit{Listing 4. Fixed Vue code.}
\end{center}
\vspace{-0.5em}
\begin{lstlisting}[numbers=left, xleftmargin=1.5em,
  basicstyle=\small\ttfamily, breaklines=true, frame=single,
  label={lst:vue_fixed},
  language=Java,
  morekeywords={import,from,export,const,defineProps},
  commentstyle=\color{gray},
  keywordstyle=\color{blue},
  stringstyle=\color{orange!70!black}]
const ResourceCard = (props: {
  iconName: string;
  title: string;
  description: string;
}) => (
  <div class="bg-white shadow rounded-lg p-6">
    <i class={`fas fa-${props.iconName} text-blue-500`}></i>
    <h2 class="text-lg font-medium">{props.title}</h2>
    <p class="text-sm text-gray-500">{props.description}</p>
  </div>
);

<div class="grid grid-cols-1 md:grid-cols-3 gap-6">
  <ResourceCard iconName="file-text" // Fixed
                title="Blog"
                description="Explore our learnings..." />
  <ResourceCard iconName="clipboard-check" // Fixed
                title="Changelog"
                description="Stay updated..." />
  <ResourceCard iconName="users" // Fixed
                title="Community"
                description="Join the discussion..." />
</div>
\end{lstlisting}

\newpage
\begin{center}
\textit{Listing 5. Wrong Angular code ($\mathcal{C}_\text{fw}$).}
\end{center}
\vspace{-0.5em}
\begin{lstlisting}[numbers=left, xleftmargin=1.5em,
  basicstyle=\small\ttfamily, breaklines=true, frame=single,
  label={lst:angular_wrong},
  language=Java,
  morekeywords={import,from,export,class,const,let},
  commentstyle=\color{gray},
  keywordstyle=\color{blue},
  stringstyle=\color{orange!70!black}]
import { NgModule } from '@angular/core';
import { CommonModule } from '@angular/common';
import { IotDashboardComponent } from
  './iot-dashboard/iot-dashboard.component';
// Missing: TemperatureDraggerComponent
// Missing: RoomSelectorComponent

@NgModule({
  declarations: [
    IotDashboardComponent,
    // TemperatureDraggerComponent removed by C_fw
    // RoomSelectorComponent removed by C_fw
  ],
  imports: [
    CommonModule,
    // NbCardModule removed by C_fw
    // NbIconModule removed by C_fw
    // NbTabsetModule removed by C_fw
  ],
})
export class DashboardModule { }
\end{lstlisting}

\begin{center}
\textit{Listing 6. Fixed Angular code.}
\end{center}
\vspace{-0.5em}
\begin{lstlisting}[numbers=left, xleftmargin=1.5em,
  basicstyle=\small\ttfamily, breaklines=true, frame=single,
  label={lst:angular_fixed},
  language=Java,
  morekeywords={import,from,export,class,const,let},
  commentstyle=\color{gray},
  keywordstyle=\color{blue},
  stringstyle=\color{orange!70!black}]
import { NgModule } from '@angular/core';
import { CommonModule } from '@angular/common';
import { NbCardModule, NbIconModule,
         NbTabsetModule } from '@nebular/theme'; // Fixed
import { IotDashboardComponent } from
  './iot-dashboard/iot-dashboard.component';
import { TemperatureDraggerComponent } from
  './iot-dashboard/temperature-dragger/
  temperature-dragger.component';           // Fixed
import { RoomSelectorComponent } from
  './iot-dashboard/room-selector/
  room-selector.component';                 // Fixed

@NgModule({
  declarations: [
    IotDashboardComponent,
    TemperatureDraggerComponent,  // Fixed
    RoomSelectorComponent,        // Fixed
  ],
  imports: [
    CommonModule,
    NbCardModule,    // Fixed
    NbIconModule,    // Fixed
    NbTabsetModule,  // Fixed
  ],
})
export class DashboardModule { }
\end{lstlisting}

\begin{figure}[!t]
  \centering
  \begin{subfigure}[b]{0.3\linewidth}
    \centering
    \includegraphics[width=\linewidth]{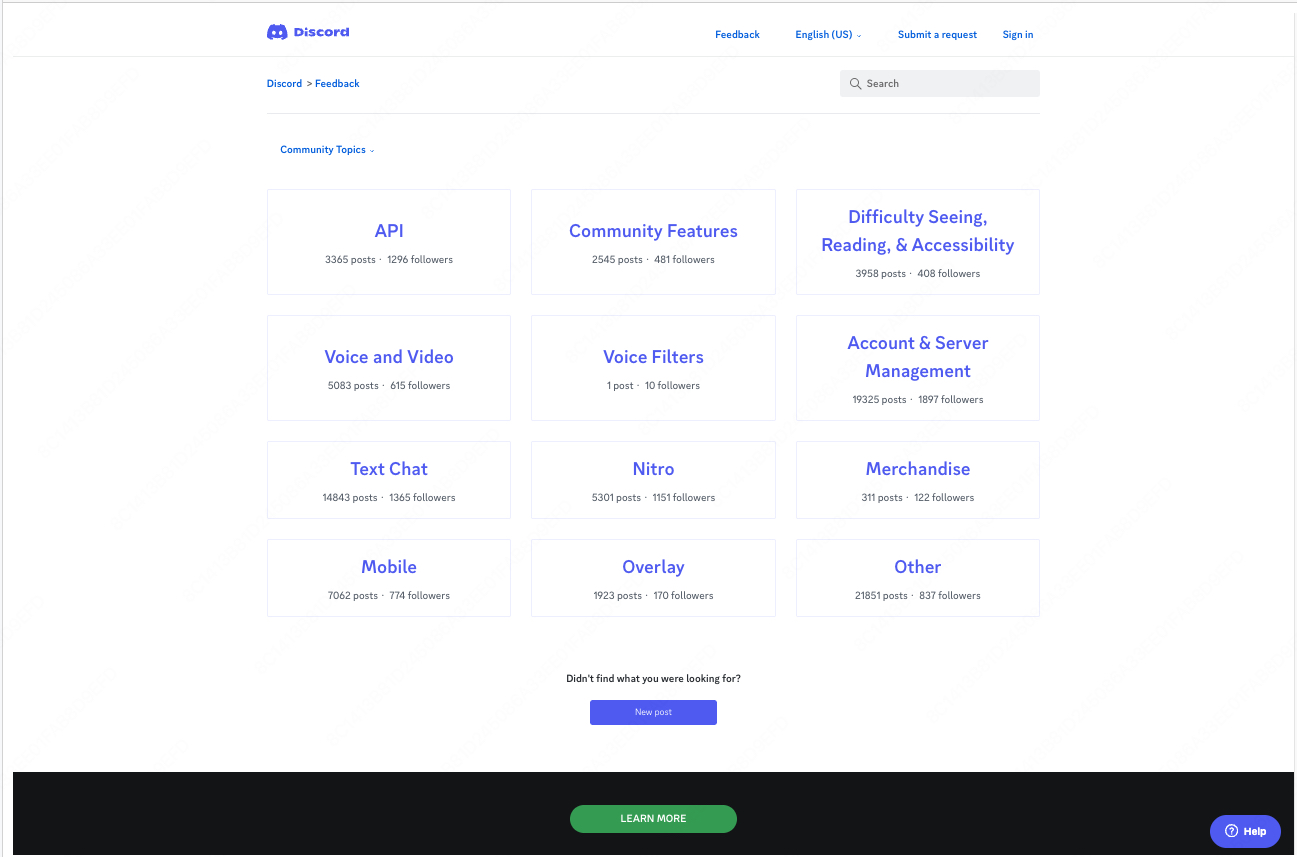}
    \caption{Original}
  \end{subfigure}
  \quad
  \begin{subfigure}[b]{0.3\linewidth}
    \centering
    \includegraphics[width=\linewidth]{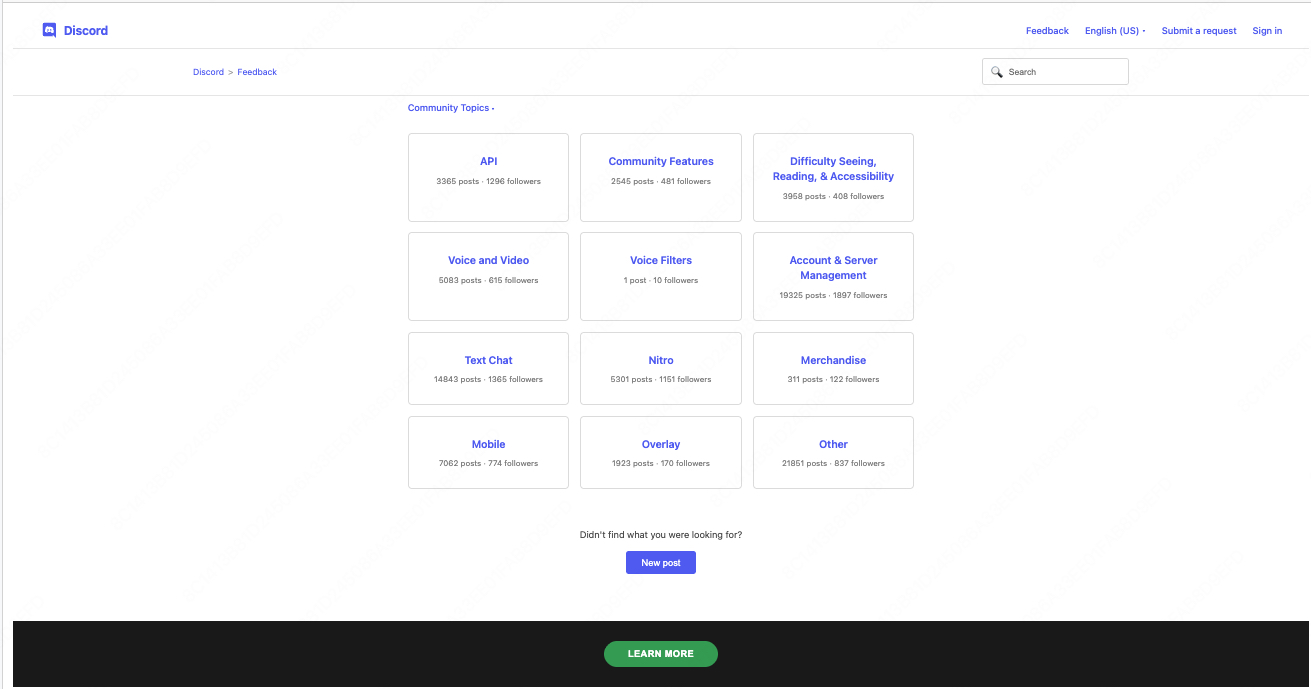}
    \caption{HTML}
  \end{subfigure}

  \vspace{0.8em}

  \begin{subfigure}[b]{0.3\linewidth}
    \centering
    \includegraphics[width=\linewidth]{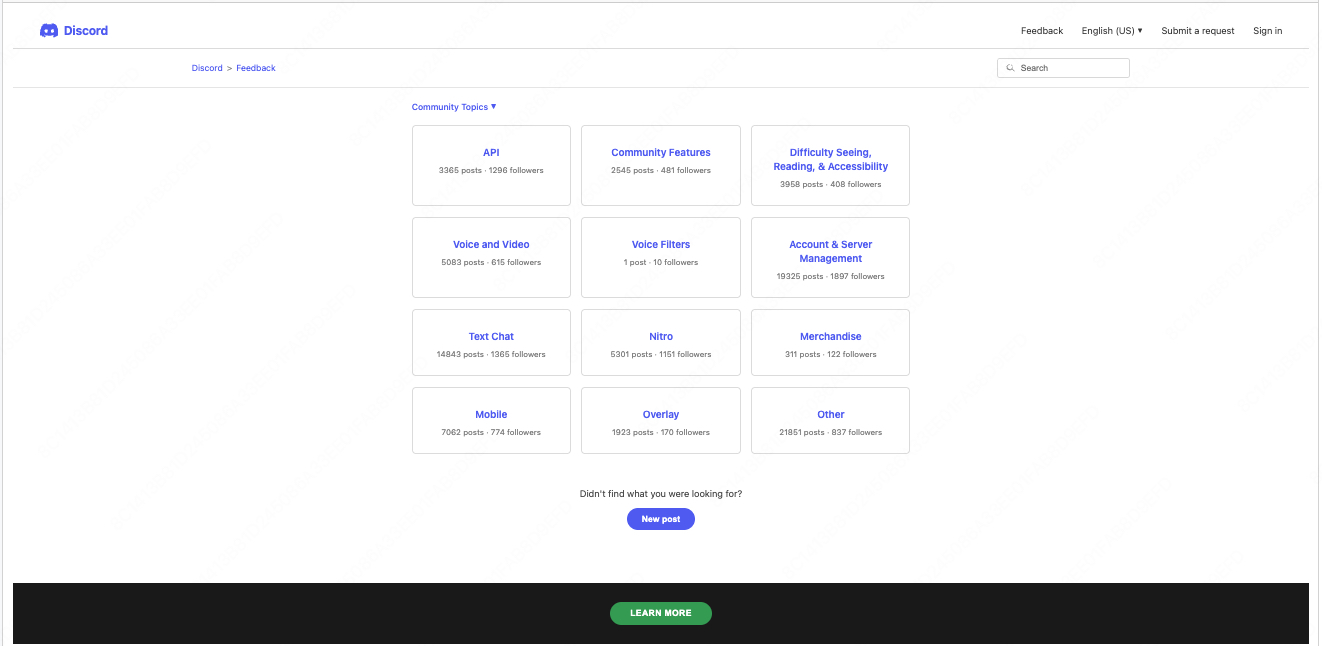}
    \caption{React}
  \end{subfigure}
  \hfill
  \begin{subfigure}[b]{0.3\linewidth}
    \centering
    \includegraphics[width=\linewidth]{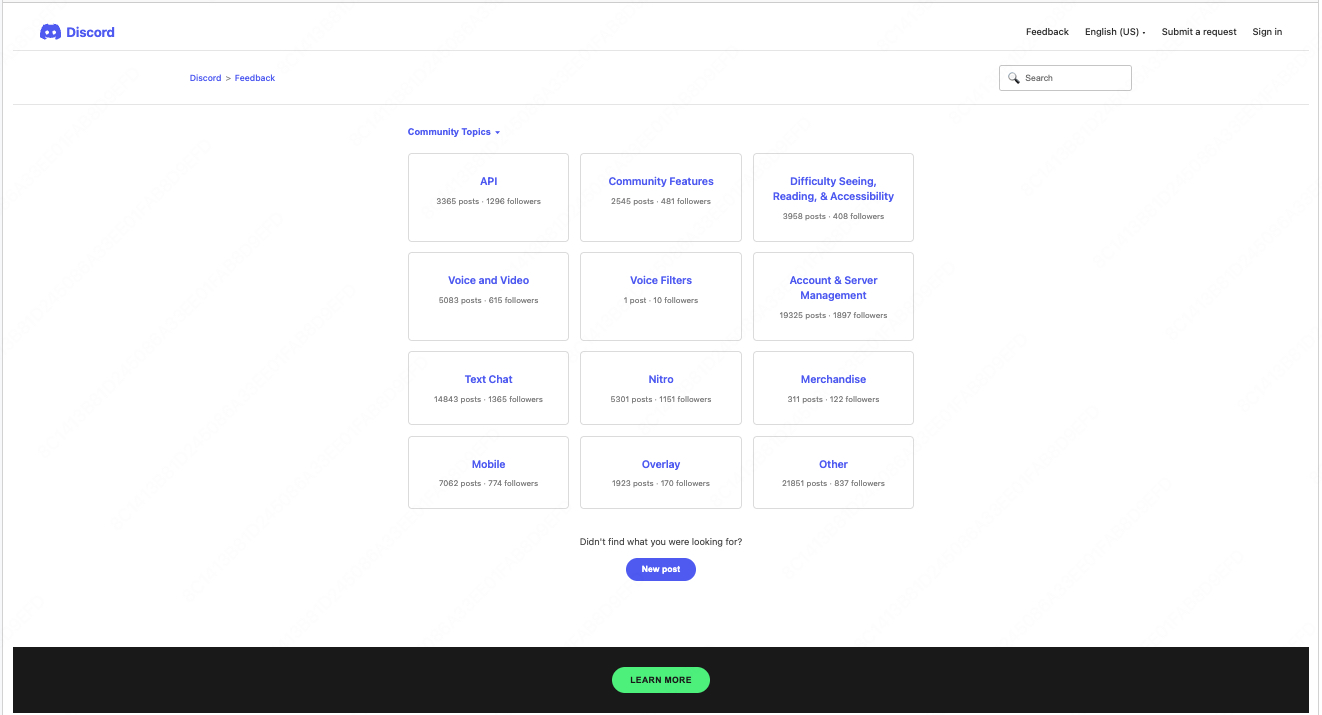}
    \caption{Vue}
  \end{subfigure}
  \hfill
  \begin{subfigure}[b]{0.3\linewidth}
    \centering
    \includegraphics[width=\linewidth]{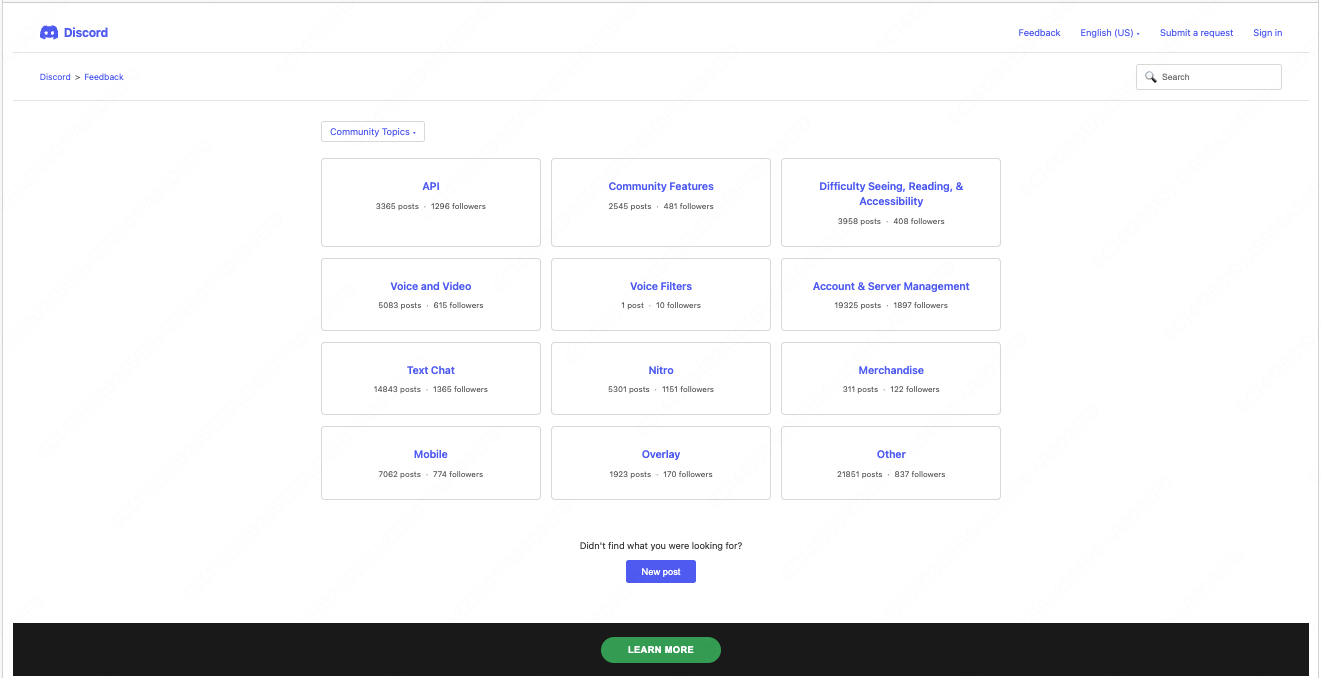}
    \caption{Angular}
  \end{subfigure}

  \vspace{1.5em}

  \begin{subfigure}[b]{0.3\linewidth}
    \centering
    \includegraphics[width=\linewidth]{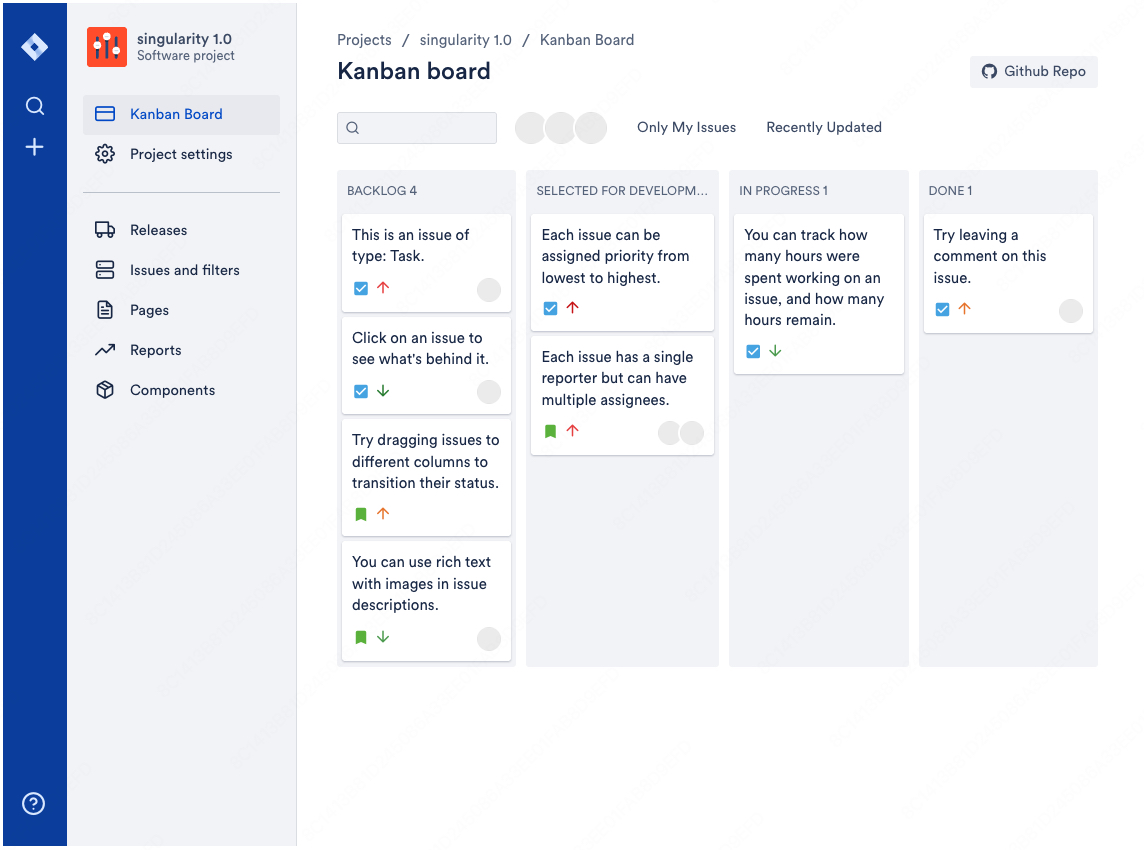}
    \caption{Original}
  \end{subfigure}
  \quad
  \begin{subfigure}[b]{0.3\linewidth}
    \centering
    \includegraphics[width=\linewidth]{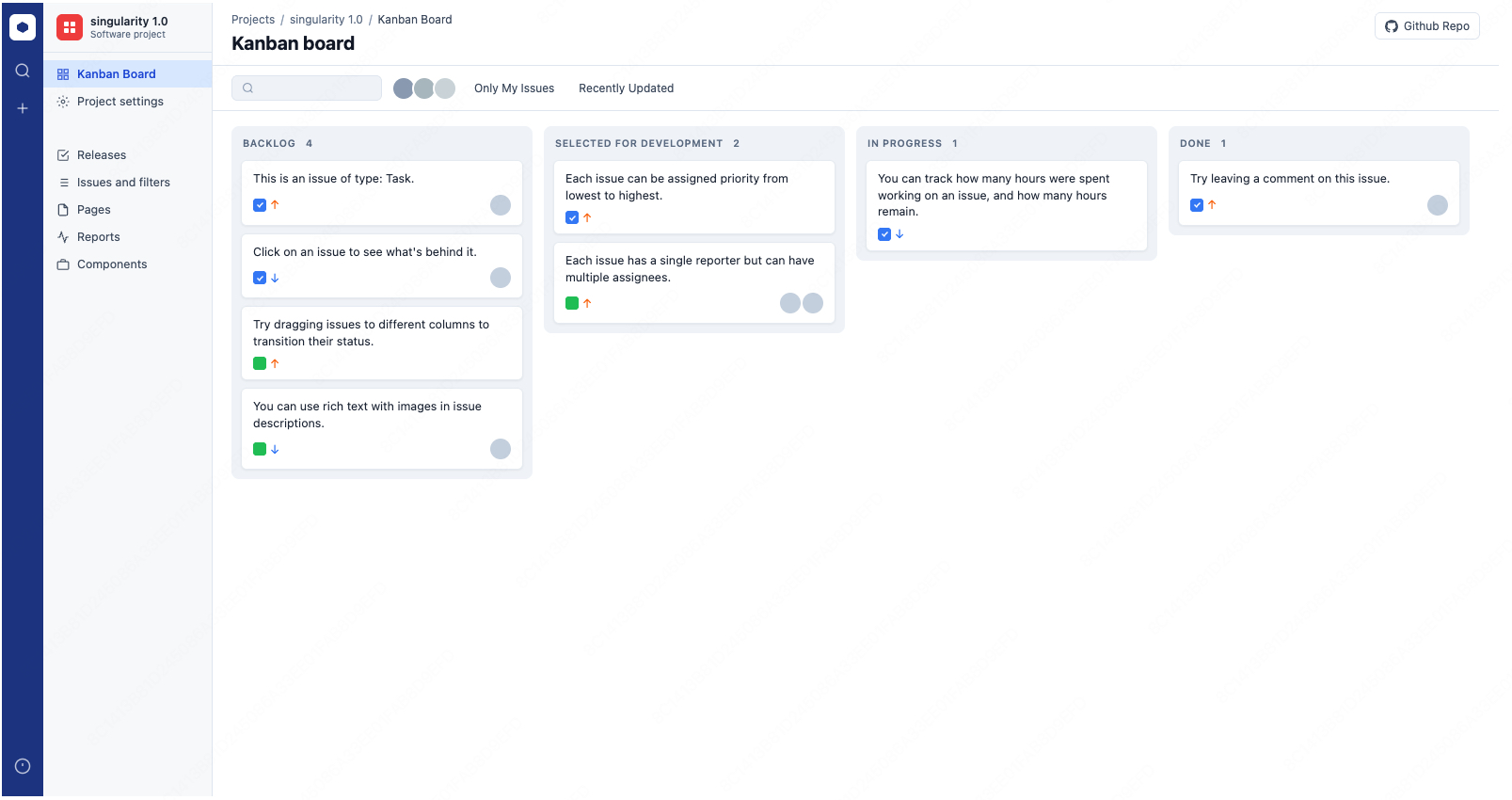}
    \caption{HTML}
  \end{subfigure}

  \vspace{0.8em}

  \begin{subfigure}[b]{0.3\linewidth}
    \centering
    \includegraphics[width=\linewidth]{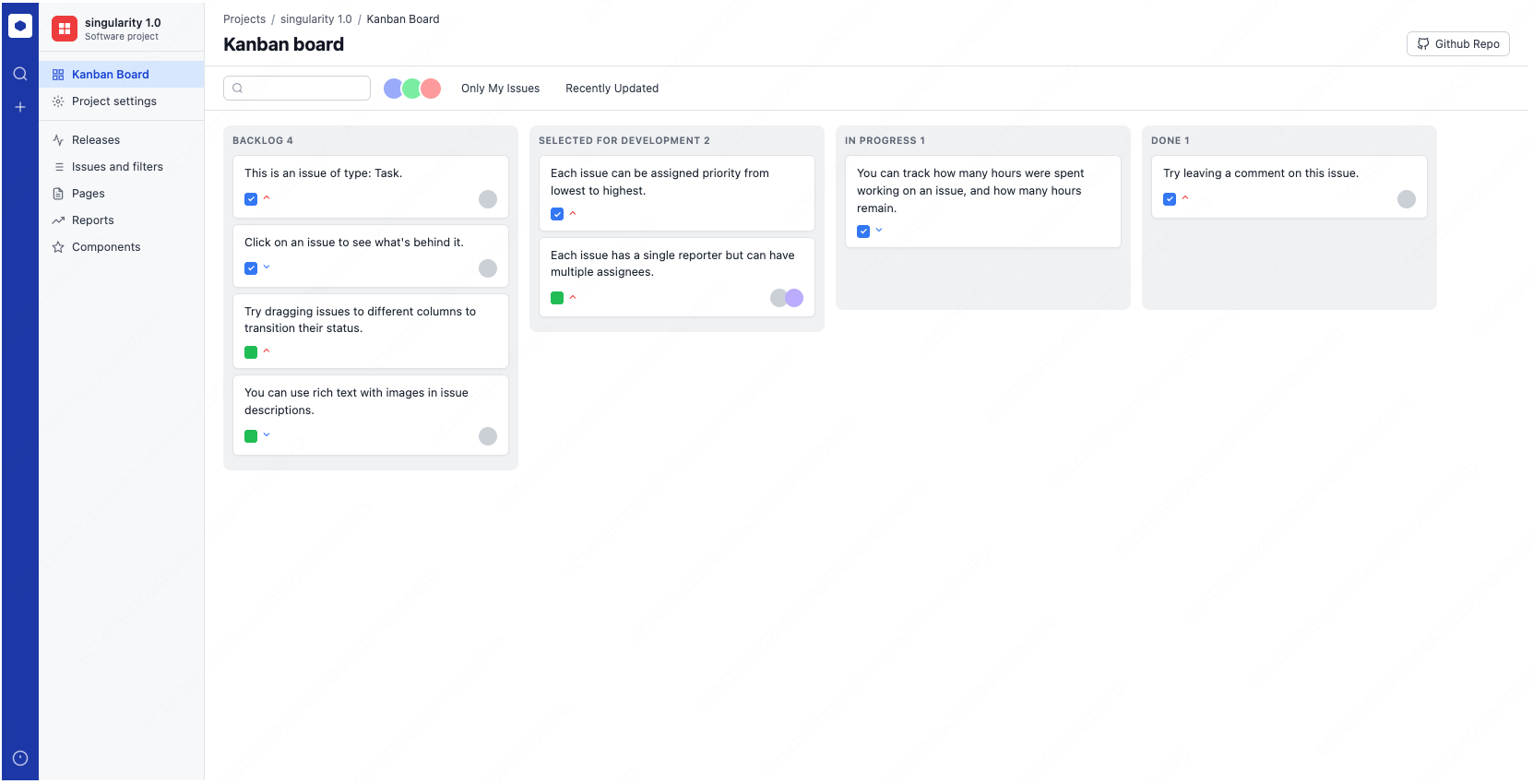}
    \caption{React}
  \end{subfigure}
  \hfill
  \begin{subfigure}[b]{0.3\linewidth}
    \centering
    \includegraphics[width=\linewidth]{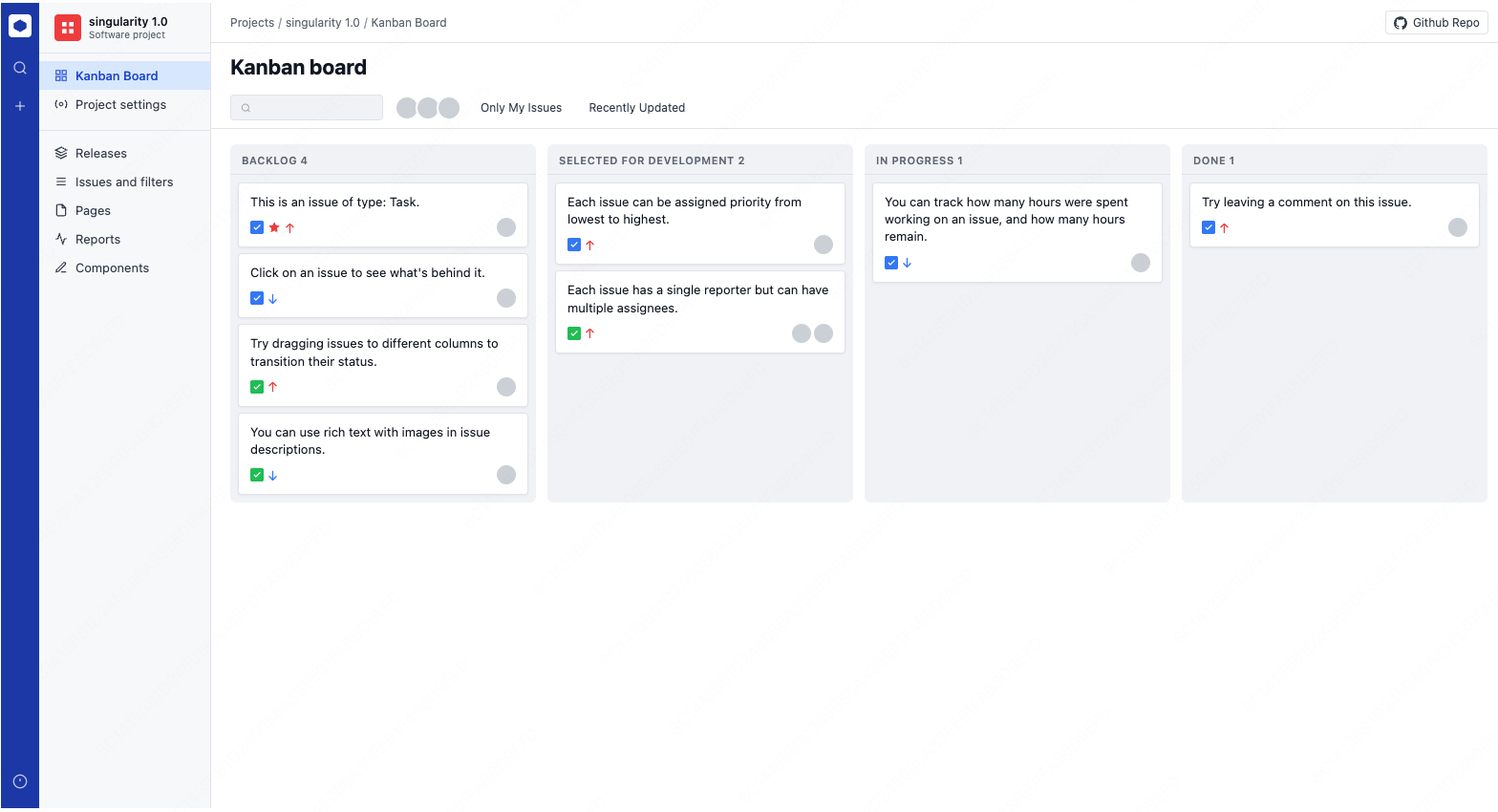}
    \caption{Vue}
  \end{subfigure}
  \hfill
  \begin{subfigure}[b]{0.3\linewidth}
    \centering
    \includegraphics[width=\linewidth]{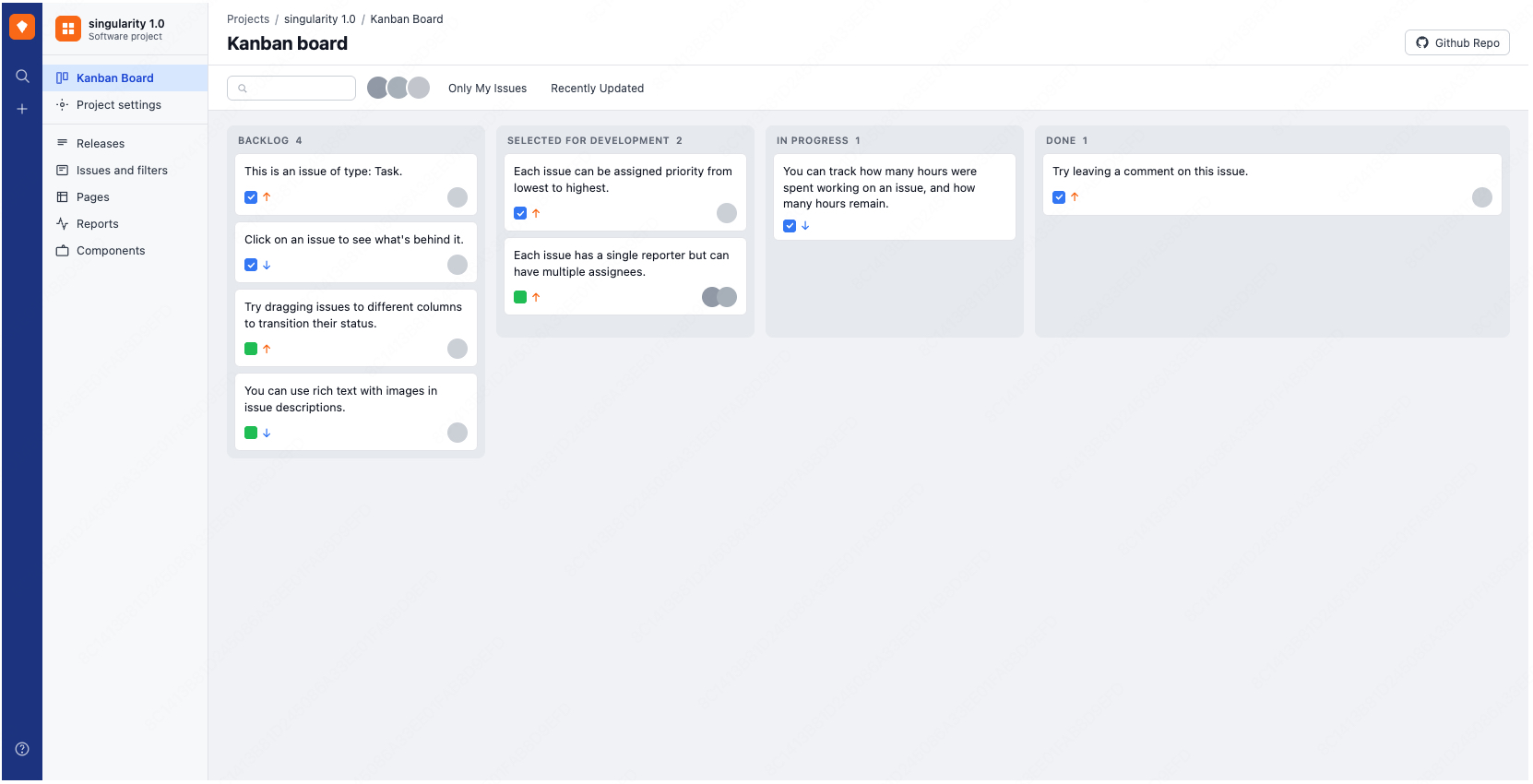}
    \caption{Angular}
  \end{subfigure}

  \vspace{1.5em}

  \begin{subfigure}[b]{0.3\linewidth}
    \centering
    \includegraphics[width=\linewidth]{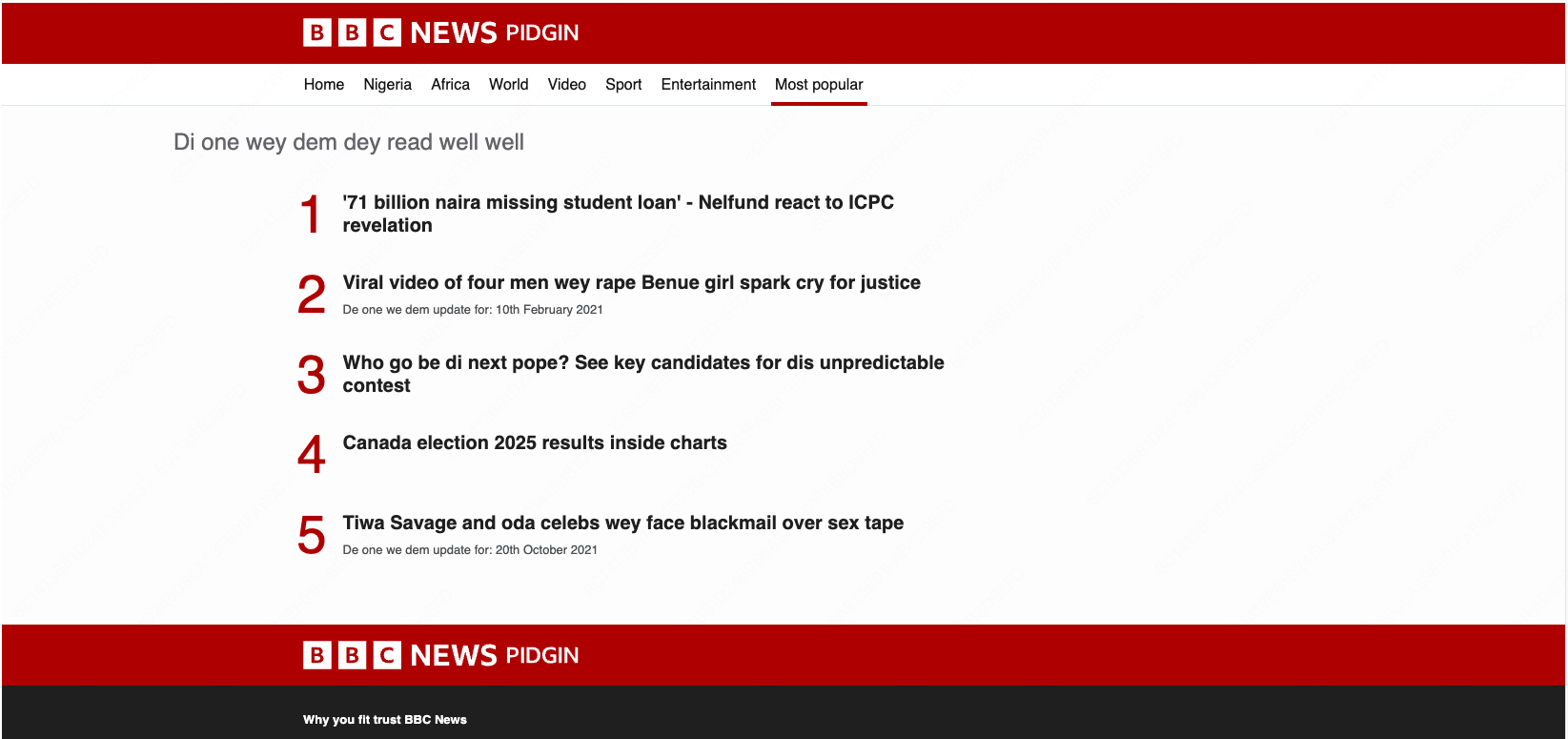}
    \caption{Original}
  \end{subfigure}
  \quad
  \begin{subfigure}[b]{0.3\linewidth}
    \centering
    \includegraphics[width=\linewidth]{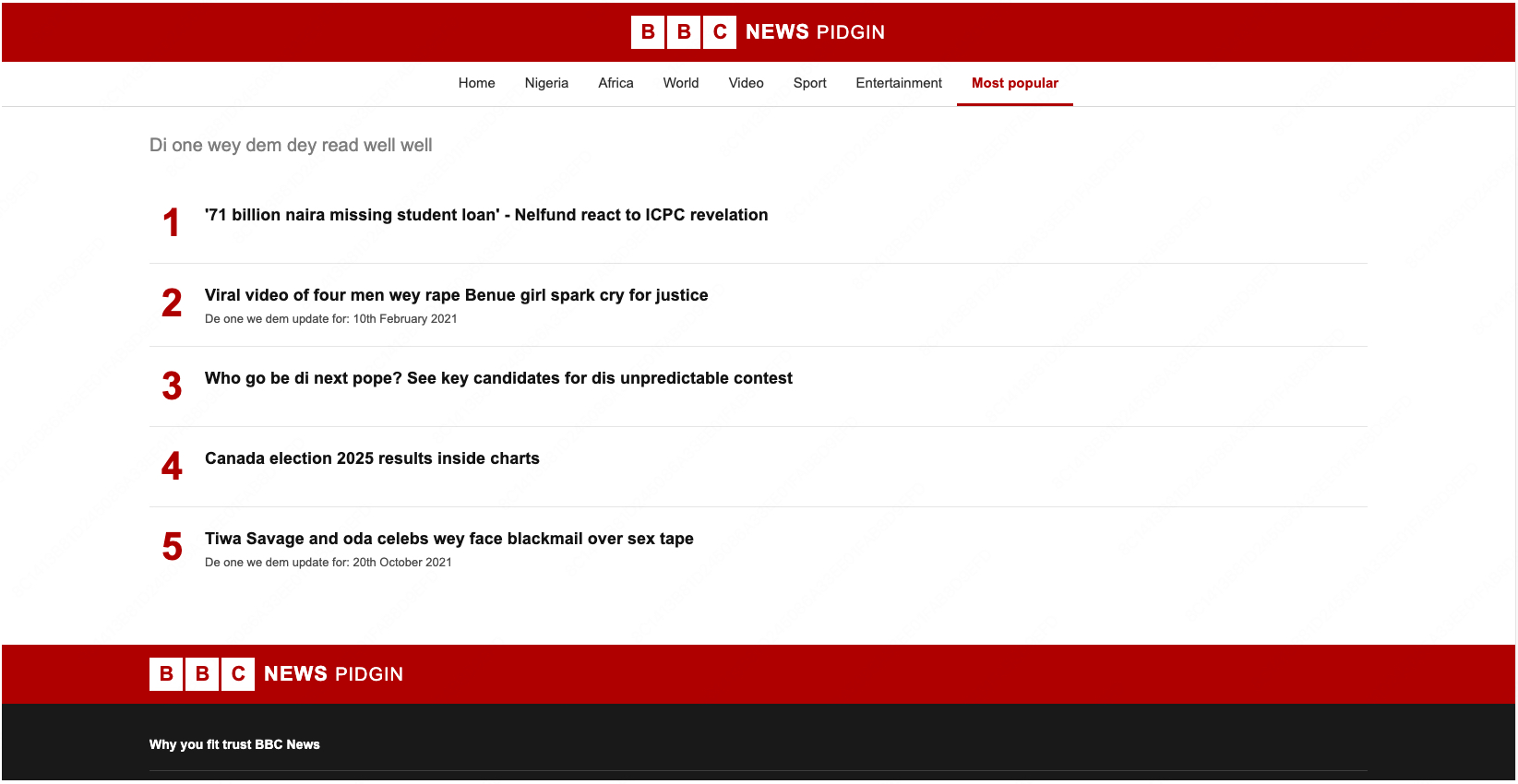}
    \caption{HTML}
  \end{subfigure}

  \vspace{0.8em}

  \begin{subfigure}[b]{0.3\linewidth}
    \centering
    \includegraphics[width=\linewidth]{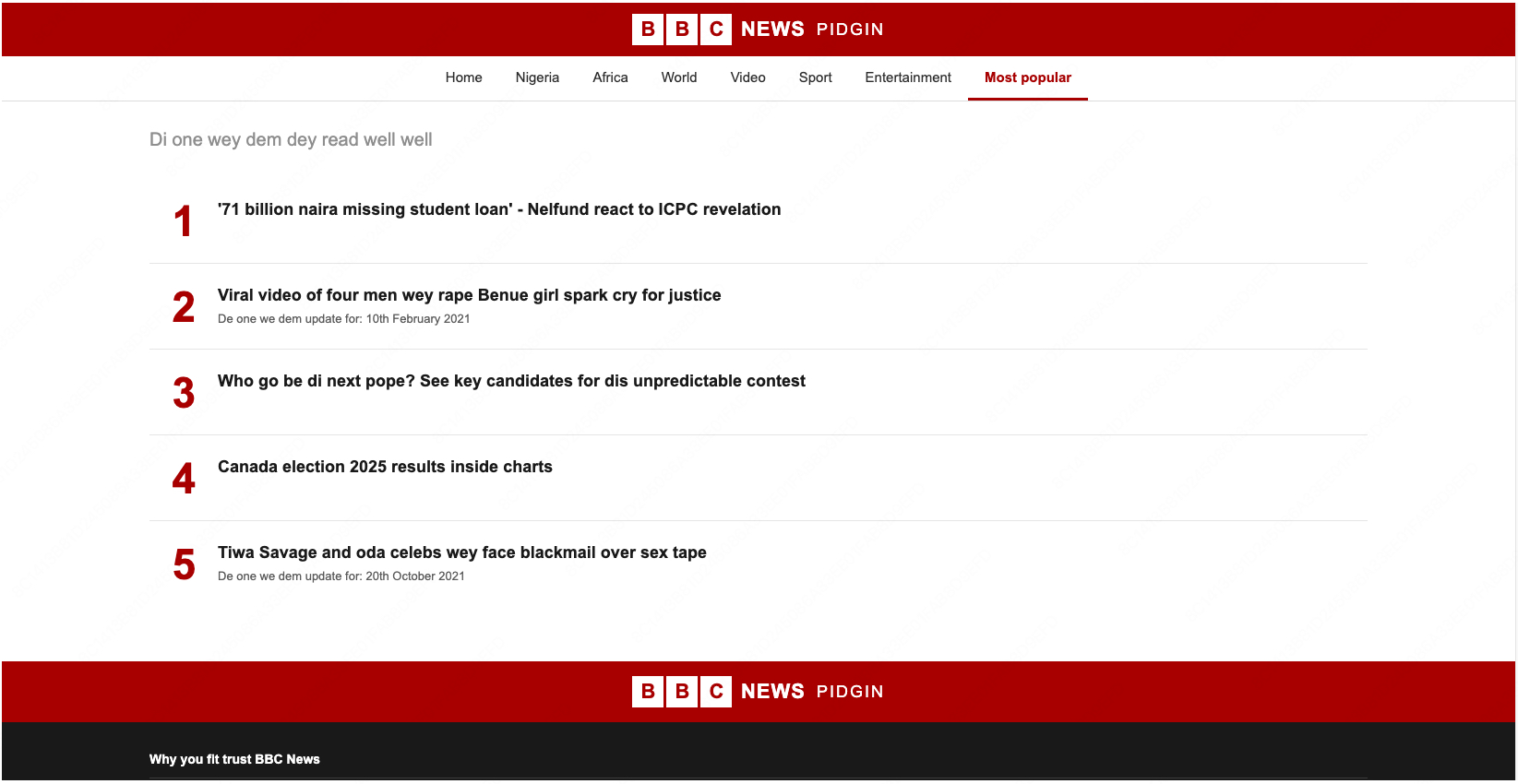}
    \caption{React}
  \end{subfigure}
  \hfill
  \begin{subfigure}[b]{0.3\linewidth}
    \centering
    \includegraphics[width=\linewidth]{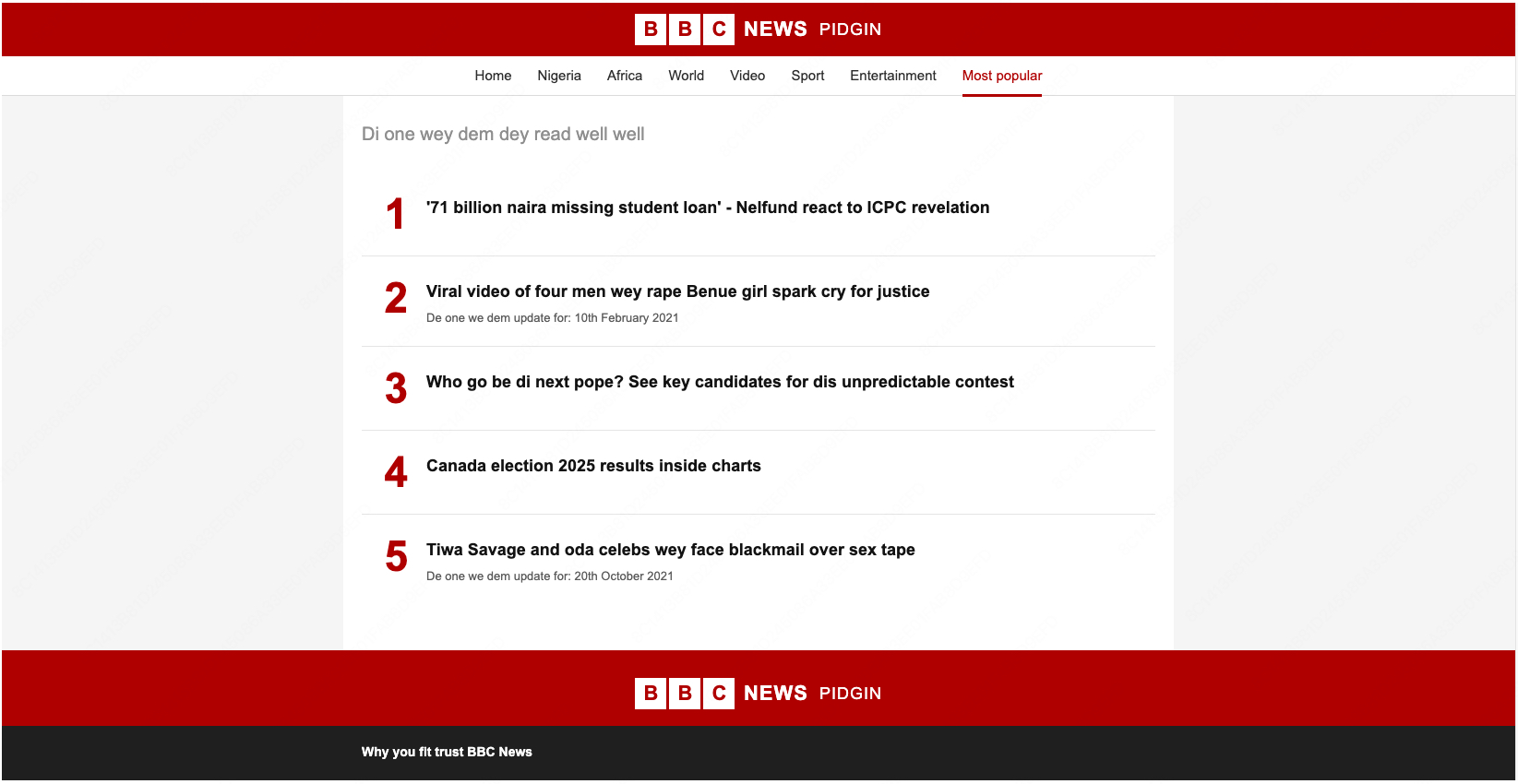}
    \caption{Vue}
  \end{subfigure}
  \hfill
  \begin{subfigure}[b]{0.3\linewidth}
    \centering
    \includegraphics[width=\linewidth]{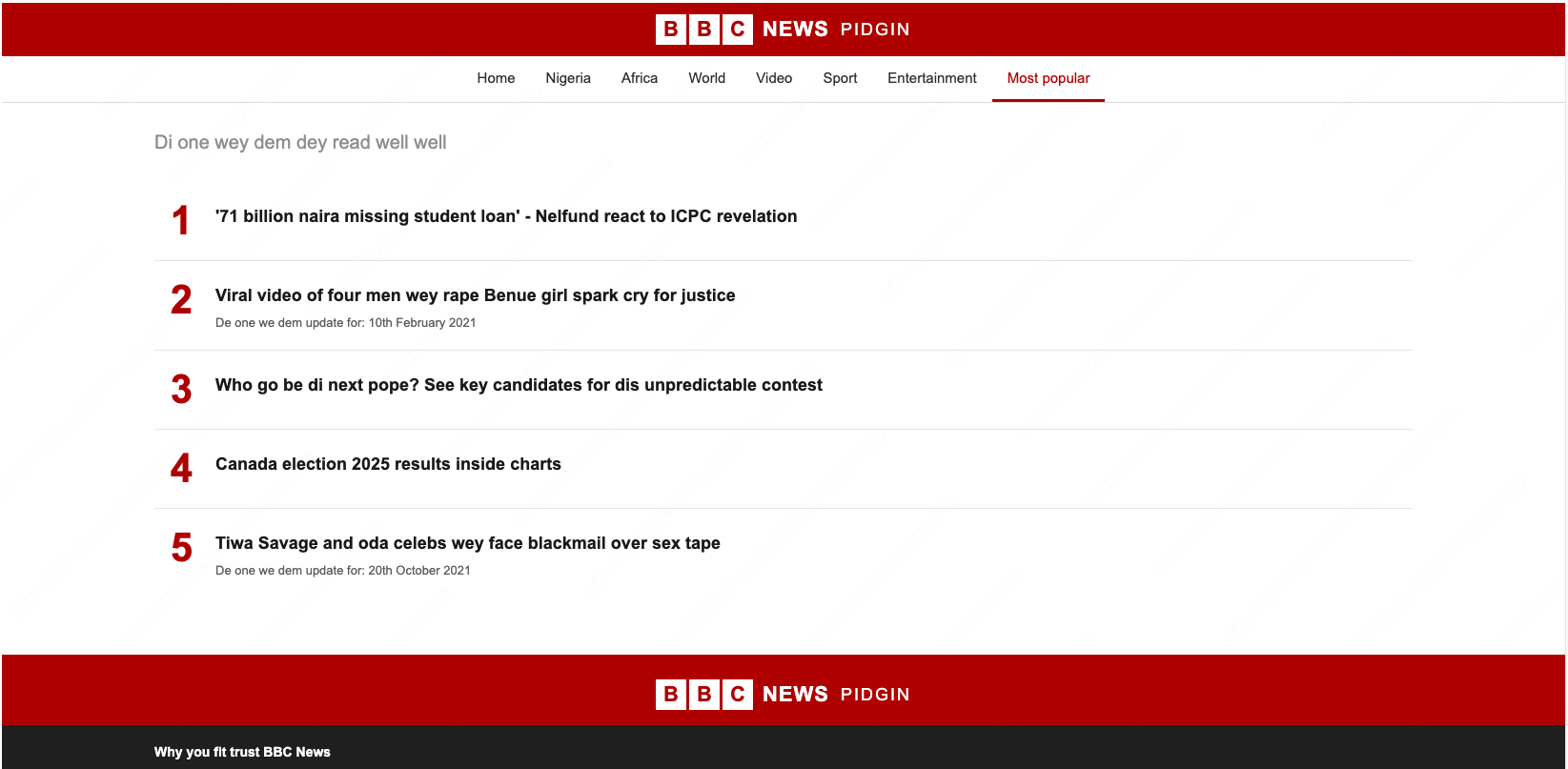}
    \caption{Angular}
  \end{subfigure}

  \caption{Qualitative rendering results of MESCoder across three UI test cases. For each case, we show the original reference screenshot (Original), an HTML generation baseline (HTML), and MESCoder outputs for three component-based frameworks (React, Vue, Angular). MESCoder faithfully reproduces the reference layout across all target frameworks.}
  \label{fig:render_compare}
\end{figure}

\paragraph{Cross-framework rendering comparison.}
Figure~\ref{fig:render_compare} shows qualitative rendering results for the same input UI across HTML, React, Vue, and Angular, alongside the original reference screenshot. The HTML output serves as a generation baseline, while MESCoder outputs for the three framework targets consistently match the reference layout.

\newpage
\section{LLM Executor Prompt Templates}
\label{app:prompts}

This appendix provides the complete prompt templates used for each action category in the MESCoder executor interface.
All prompts share a common system preamble that establishes the task context; per-action sections then specify the concrete instruction, the project context, and the expected output format.
Identical templates are used across all six LLM executors; no executor-specific prompt engineering is applied.
Variables enclosed in $\langle\rangle$ are filled by the framework at inference time.

\subsection{System Preamble (Shared Across All Actions)}

The following system message is prepended to every LLM call:

\begin{quote}
\small\ttfamily
You are a senior front-end engineer specializing in $\langle$\textit{framework}$\rangle$ (one of React, Vue, Angular).
You are assisting an automated code-generation agent that constructs a multi-file
$\langle$\textit{framework}$\rangle$ project from a UI screenshot.
The project is built incrementally: at each step you receive a partial project state,
the current compilation errors (if any), and an action instruction.
Your response must follow the structured output format exactly:
produce only the requested code edits, with no surrounding prose, no markdown
code fences outside the prescribed format, and no commentary.
Each edit is a JSON object: \{``file'': ``<path>'', ``old'': ``<old\_content>'',
``new'': ``<new\_content>''\}.
If you are generating a new file, set ``old'' to an empty string.
\end{quote}

\subsection{Continuation Prompt (\textsc{Continue} Action)}

The \textsc{Continue} action advances the current task when the partial project is sufficiently healthy, generating code for the next scaffold node.
The prompt is structured as follows:

\begin{quote}
\small\ttfamily
\textbf{[TASK]} Generate the $\langle$\textit{framework}$\rangle$ component for the UI element described below.\\
\\
\textbf{[SCAFFOLD NODE]}\\
- Type: $\langle$\textit{node\_type}$\rangle$ (e.g., ``Card'', ``NavBar'', ``DataTable'')\\
- Bounding box (normalized): $\langle$\textit{x\_min, y\_min, x\_max, y\_max}$\rangle$\\
- Tree depth: $\langle$\textit{depth}$\rangle$\\
- Children to generate later: $\langle$\textit{child\_count}$\rangle$\\
\\
\textbf{[CONTENT SLOTS]}\\
$\langle$\textit{slot\_table}$\rangle$
(Each slot: \{node\_id, text\_content, img\_crop\_description\})\\
\\
\textbf{[SCREENSHOT DESCRIPTION]}\\
$\langle$\textit{screenshot image or visual description appended here}$\rangle$\\
\\
\textbf{[EXISTING PROJECT FILES]}\\
$\langle$\textit{list of already-generated files with their current content}$\rangle$\\
\\
\textbf{[INSTRUCTION]}\\
Generate the file \texttt{$\langle$\textit{target\_filename}$\rangle$} for this component.
Implement the visual layout faithfully.
Use only $\langle$\textit{framework}$\rangle$-idiomatic patterns.
Do not import any external UI libraries unless they are already present in package.json.
Declare all props with correct TypeScript types.\\
\\
\textbf{[OUTPUT FORMAT]}\\
\{"file": "$\langle$\textit{target\_filename}$\rangle$", "old": "", "new": "<complete file content>"\}
\end{quote}

\subsection{Assembly Prompt (\textsc{Assemble} Action)}

The \textsc{Assemble} action integrates completed child units into the current parent node, resolving imports and type compatibility.

\begin{quote}
\small\ttfamily
\textbf{[TASK]} Assemble the parent component \texttt{$\langle$\textit{parent\_filename}$\rangle$} by integrating the
following child components that have already been generated.\\
\\
\textbf{[PARENT NODE]}\\
- Type: $\langle$\textit{parent\_node\_type}$\rangle$\\
- Expected children: $\langle$\textit{list of child component names and filenames}$\rangle$\\
\\
\textbf{[CHILD COMPONENT INTERFACES]}\\
$\langle$\textit{For each child: filename, exported component name, prop interface definition}$\rangle$\\
\\
\textbf{[EXISTING PARENT FILE (if any)]}\\
$\langle$\textit{current content of the parent file, or empty string if new}$\rangle$\\
\\
\textbf{[CURRENT DIAGNOSTIC]}\\
$\langle$\textit{sandbox error messages for the current scope, or ``No errors'' if clean}$\rangle$\\
\\
\textbf{[INSTRUCTION]}\\
Write the complete \texttt{$\langle$\textit{parent\_filename}$\rangle$} that:
(1) imports all child components with correct relative paths;
(2) uses each child component with correct props in the parent JSX/template;
(3) passes no incorrect prop types and omits no required props;
(4) is compatible with the framework's assembly conventions
    (e.g., Angular \texttt{@NgModule} declarations, Vue \texttt{components} option,
     React named imports).\\
\\
\textbf{[OUTPUT FORMAT]}\\
\{"file": "$\langle$\textit{parent\_filename}$\rangle$", "old": "$\langle$\textit{old content or empty}$\rangle$", "new": "$\langle$\textit{new content}$\rangle$"\}
\end{quote}

\subsection{Local Repair Prompt (\textsc{PatchLocal} Action)}

The \textsc{PatchLocal} action invokes a local repair over a small scope, targeting a single file identified as the primary error source.

\begin{quote}
\small\ttfamily
\textbf{[TASK]} Fix the compilation error in a single file.\\
\\
\textbf{[ERROR SUMMARY]}\\
$\langle$\textit{diagnostic summary: dominant error family, error messages, line numbers}$\rangle$\\
\\
\textbf{[TARGET FILE]}\\
Filename: \texttt{$\langle$\textit{target\_filename}$\rangle$}\\
$\langle$\textit{full file content}$\rangle$\\
\\
\textbf{[DIRECT IMPORT DEPENDENTS (read-only context)]}\\
$\langle$\textit{for each file that imports target\_filename: filename and content}$\rangle$\\
\\
\textbf{[CONSTRAINT]}\\
You may ONLY edit \texttt{$\langle$\textit{target\_filename}$\rangle$}.
Do NOT modify any dependent files.
Do NOT change the component's exported interface (prop names, types, return shape)
unless the error explicitly requires it.\\
\\
\textbf{[INSTRUCTION]}\\
Produce a minimal, correct fix for the listed errors.
Preserve all existing functionality outside the error site.\\
\\
\textbf{[OUTPUT FORMAT]}\\
\{"file": "$\langle$\textit{target\_filename}$\rangle$", "old": "$\langle$\textit{exact old content}$\rangle$", "new": "$\langle$\textit{corrected content}$\rangle$"\}
\end{quote}

\subsection{Cluster Repair Prompt (\textsc{PatchCluster} Action)}

The \textsc{PatchCluster} action invokes a broader repair over a small dependency or assembly cluster identified by the sandbox.

\begin{quote}
\small\ttfamily
\textbf{[TASK]} Fix a cross-file dependency error spanning multiple files.\\
\\
\textbf{[ERROR CLUSTER SUMMARY]}\\
Error family: $\langle$\textit{one of: Assembly \& Hierarchy; Module Dep.\@ \& Import/Export; Framework-Specific Engineering}$\rangle$\\
Affected files: $\langle$\textit{list of filenames in the dependency cluster}$\rangle$\\
Error messages:\\
$\langle$\textit{full compiler/runtime error output for the cluster}$\rangle$\\
\\
\textbf{[CLUSTER FILE CONTENTS]}\\
$\langle$\textit{for each file in the cluster: filename and full content}$\rangle$\\
\\
\textbf{[INSTRUCTION]}\\
Produce a coordinated set of edits that resolves ALL listed errors simultaneously.
You MAY edit any subset of the files in the cluster.
Ensure that after your edits: (1) no new errors are introduced in files outside
the cluster, (2) previously passing files in the project are not broken,
(3) the overall UI layout and component hierarchy is preserved.\\
\\
\textbf{[OUTPUT FORMAT]}\\
Return a JSON array of edit objects, one per modified file:\\
{[}\{"file": "$\langle$\textit{filename\_1}$\rangle$", "old": "$\langle$\textit{old}$\rangle$", "new": "$\langle$\textit{new}$\rangle$"\},\\
\quad\{"file": "$\langle$\textit{filename\_2}$\rangle$", "old": "$\langle$\textit{old}$\rangle$", "new": "$\langle$\textit{new}$\rangle$"\},\\
\quad\ldots{}{]}\\
\end{quote}

\subsection{Finish Prompt (\textsc{Finish} Action)}

The \textsc{Finish} action terminates the episode when the project is complete and executable, applying any remaining fixes in a final verification pass.

\begin{quote}
\small\ttfamily
\textbf{[TASK]} Perform a final review and apply any remaining fixes to ensure the project
compiles and renders correctly as a whole.\\
\\
\textbf{[FULL PROJECT STATE]}\\
$\langle$\textit{all project files listed with their content}$\rangle$\\
\\
\textbf{[FINAL DIAGNOSTIC]}\\
$\langle$\textit{output of full-project verification (tsc / vue-tsc / ng build); or ``All checks pass''}$\rangle$\\
\\
\textbf{[INSTRUCTION]}\\
If the final diagnostic reports errors, fix them with minimal changes.
If the diagnostic is clean, verify that:
(1) \texttt{package.json} contains all required dependencies;
(2) the entry point (e.g., \texttt{src/main.tsx}, \texttt{src/main.ts}) correctly
    mounts the root component;
(3) all image or asset references resolve to valid paths.
Return only the files that require changes; return an empty array if the project
is already complete and correct.\\
\\
\textbf{[OUTPUT FORMAT]}\\
JSON array of edit objects (may be empty \texttt{[]} if no changes needed).\\
\end{quote}

\section{Limitations}
\label{app:Limitations}

Our framework improves executable multi-framework front-end generation through execution feedback and self-correction, but it still has several limitations. First, the system depends on the quality of the scaffold preprocessor and the runtime sandbox. Incorrect structure extraction or incomplete framework-specific checks may mislead the policy and the LLM executor. Second, although the policy learns high-level corrective decisions, the actual code edits are still produced by the LLM executor; when the executor repeatedly fails to generate valid patches, the policy alone cannot fully recover the project. Third, online reinforcement learning introduces additional computational cost due to repeated LLM and sandbox calls, which may limit deployment in latency-sensitive settings. Finally, our method focuses on syntactic, structural, and build-time executability, while broader software engineering aspects such as security, accessibility, maintainability, and long-term code quality are not fully addressed.

\section{Broader Impacts}
\label{app:Impacts}

This work may have positive societal impacts by reducing the cost of front-end development, enabling faster prototyping, and making UI implementation more accessible to designers, small teams, and developers with limited framework expertise. By explicitly emphasizing executable multi-framework projects, the method may also reduce debugging burden and improve the reliability of AI-assisted software generation. At the same time, the work has potential negative impacts. More capable automated code generation could contribute to low-quality mass-produced web interfaces, overreliance on generated code, or deployment of projects whose security, accessibility, or privacy properties have not been adequately reviewed. It may also affect demand for certain routine front-end implementation tasks. Therefore, generated projects should be treated as developer-assistance outputs rather than fully autonomous production-ready systems, and should undergo human review, security checks, and accessibility validation before deployment.

\newpage
\section*{NeurIPS Paper Checklist}

\begin{enumerate}

\item {\bf Claims}
    \item[] Question: Do the main claims made in the abstract and introduction accurately reflect the paper's contributions and scope?
    \item[] Answer: \answerYes{}{} %
    \item[] Justification: The abstract and introduction state the main scope, method, and empirical claims.

\item {\bf Limitations}
    \item[] Question: Does the paper discuss the limitations of the work performed by the authors?
    \item[] Answer: \answerYes{} %
    \item[] Justification: We have dicussed three limitations in Appendix~\ref{app:Limitations}.

\item {\bf Theory assumptions and proofs}
    \item[] Question: For each theoretical result, does the paper provide the full set of assumptions and a complete (and correct) proof?
    \item[] Answer: \answerNA{} %
    \item[] Justification: This paper does not include formal theoretical results or proofs.

    \item {\bf Experimental result reproducibility}
    \item[] Question: Does the paper fully disclose all the information needed to reproduce the main experimental results of the paper to the extent that it affects the main claims and/or conclusions of the paper (regardless of whether the code and data are provided or not)?
    \item[] Answer: \answerYes{} %
    \item[] Justification: See Section~\ref{sec:setup}, and Appendix~\ref{app:sandbox_data}.

\item {\bf Open access to data and code}
    \item[] Question: Does the paper provide open access to the data and code, with sufficient instructions to faithfully reproduce the main experimental results, as described in supplemental material?
    \item[] Answer: \answerYes{} %
    \item[] Justification: Our code is publicly available.

\item {\bf Experimental setting/details}
    \item[] Question: Does the paper specify all the training and test details (e.g., data splits, hyperparameters, how they were chosen, type of optimizer) necessary to understand the results?
    \item[] Answer: \answerYes{} %
    \item[] Justification: In Section 4

\item {\bf Experiment statistical significance}
    \item[] Question: Does the paper report error bars suitably and correctly defined or other appropriate information about the statistical significance of the experiments?
    \item[] Answer: \answerYes{} %
    \item[] Justification: In Section 4

\item {\bf Experiments compute resources}
    \item[] Question: For each experiment, does the paper provide sufficient information on the computer resources (type of compute workers, memory, time of execution) needed to reproduce the experiments?
    \item[] Answer: \answerYes{} %
    \item[] Justification: See Appendix~\ref{sec:impl_details} and Appendix~\ref{app:efficiency_detail}.
    
\item {\bf Code of ethics}
    \item[] Question: Does the research conducted in the paper conform, in every respect, with the NeurIPS Code of Ethics \url{https://neurips.cc/public/EthicsGuidelines}?
    \item[] Answer: \answerYes{} %
    \item[] Justification: The work is a front-end code generation method and appears consistent with the NeurIPS Code of Ethics.

\item {\bf Broader impacts}
    \item[] Question: Does the paper discuss both potential positive societal impacts and negative societal impacts of the work performed?
    \item[] Answer: \answerYes{}.
    \item[] Justification: We discussed impacts in Appendix \ref{app:Impacts}.
    
\item {\bf Safeguards}
    \item[] Question: Does the paper describe safeguards that have been put in place for responsible release of data or models that have a high risk for misuse (e.g., pre-trained language models, image generators, or scraped datasets)?
    \item[] Answer: \answerNA{} %
    \item[] Justification:  The paper poses no such risks

\item {\bf Licenses for existing assets}
    \item[] Question: Are the creators or original owners of assets (e.g., code, data, models), used in the paper, properly credited and are the license and terms of use explicitly mentioned and properly respected?
    \item[] Answer: \answerYes{} %
    \item[] Justification: Properly cited and stated

\item {\bf New assets}
    \item[] Question: Are new assets introduced in the paper well documented and is the documentation provided alongside the assets?
    \item[] Answer: \answerYes{} %
    \item[] Justification: The code we released in this paper contains related documents.

\item {\bf Crowdsourcing and research with human subjects}
    \item[] Question: For crowdsourcing experiments and research with human subjects, does the paper include the full text of instructions given to participants and screenshots, if applicable, as well as details about compensation (if any)? 
    \item[] Answer: \answerNA{} %
    \item[] Justification: Not related

\item {\bf Institutional review board (IRB) approvals or equivalent for research with human subjects}
    \item[] Question: Does the paper describe potential risks incurred by study participants, whether such risks were disclosed to the subjects, and whether Institutional Review Board (IRB) approvals (or an equivalent approval/review based on the requirements of your country or institution) were obtained?
    \item[] Answer: \answerNA{} %
    \item[] Justification: Not related

\item {\bf Declaration of LLM usage}
    \item[] Question: Does the paper describe the usage of LLMs if it is an important, original, or non-standard component of the core methods in this research? Note that if the LLM is used only for writing, editing, or formatting purposes and does \emph{not} impact the core methodology, scientific rigor, or originality of the research, declaration is not required.
    \item[] Answer: \answerYes{} %
    \item[] Justification: As described in the methodology, MESCoder uses an action-conditioned LLM executor to generate and patch code under policy-selected actions. Any additional use of LLMs for polishing or formatting did not affect the research methods or contributions.

\end{enumerate}

\end{document}